\documentclass[12pt]{iopart}

\expandafter\let\csname equation*\endcsname\relax
\expandafter\let\csname endequation*\endcsname\relax

\usepackage{amsmath}
\usepackage{graphicx}
\usepackage{amsfonts}
\usepackage{aas_macros} 
\usepackage{booktabs}
\usepackage{caption} 

\usepackage{xcolor}

\usepackage[]{hyperref}
\hypersetup{
    colorlinks, linkcolor=red, 
    citecolor=blue,
    }

\usepackage[english]{babel}
\usepackage{csquotes} 
\addto\extrasenglish{

}

\usepackage{fancyhdr}
\usepackage{cite}
\usepackage{bm}     

\begin{document}

\title{Learning and Interpreting Gravitational-Wave Features from CNNs with a Random Forest Approach}

\author{Jun Tian}
\address{School of Physics and Astronomy, China West Normal University, Nanchong 637002, China}
\ead{tianjun@stu.cwnu.edu.cn}

\author{He Wang}
\address{International Centre for Theoretical Physics Asia-Pacific (ICTP-AP), University of Chinese Academy of Sciences (UCAS), Beijing 100049, China;}
\address{Taiji Laboratory for Gravitational Wave Universe (Beijing/Hangzhou), University of Chinese Academy of Sciences (UCAS), Beijing 100049, China}
\ead{hewang@ucas.ac.cn}

\author{Jibo He}
\address{International Centre for Theoretical Physics Asia-Pacific (ICTP-AP), University of Chinese Academy of Sciences (UCAS), Beijing 100049, China;}
\address{Taiji Laboratory for Gravitational Wave Universe (Beijing/Hangzhou), University of Chinese Academy of Sciences (UCAS), Beijing 100049, China;}
\address{Hangzhou Institute for Advanced Study, UCAS, Hangzhou 310024, China}
\ead{jibo.he@ucas.ac.cn}

\author{Yu Pan}
\address{School of Electronic Science and Engineering, Chongqing University of Posts and Telecommunications, Chongqing 400065, China}

\author{Shuo Cao}
\address{Institute for Frontiers in Astronomy and Astrophysics, Beijing Normal University, Beijing 102206, China;}
\address{School of Physics and Astronomy, Beijing Normal University, Beijing 100875, China}

\author{Qingquan Jiang}
\address{School of Physics and Astronomy, China West Normal University, Nanchong 637002, China}
\ead{qqjiangphys@yeah.net}

\vspace{10pt}

\newpage

\begin{abstract}
Convolutional neural networks (CNNs) have attracted increasing attention in gravitational wave (GW) data analysis due to their ability to automatically learn hierarchical features from raw strain data.
However, the physical meaning of these learned features remains underexplored, limiting the interpretability of such models. In this work, we propose a hybrid architecture that combines a CNN-based feature extractor with a random forest~(RF) classifier to improve both detection performance and interpretability. Unlike prior approaches that directly connect classifiers to CNN outputs, our method introduces four physically interpretable metrics—variance, signal-to-noise ratio (SNR), waveform overlap, and peak amplitude—computed from the final convolutional layer. These are jointly used with the CNN output probability in the RF classifier to enable more informed decision boundaries.
Tested on long-duration strain datasets, our hybrid model outperforms a baseline-CNN model, achieving a relative improvement of 21\% in sensitivity at a fixed false alarm rate of 10 events per month. Notably, it also shows improved detection of low-SNR signals (SNR $\le$ 10), which are especially vulnerable to misclassification in noisy environments.
Feature importance analysis and ablation studies reveal that handcrafted features play a significant role in classification decisions and contribute to improved performance.
These findings suggest that physically motivated post-processing of CNN feature maps can serve as a valuable tool for interpretable and efficient GW detection, bridging the gap between deep learning and domain knowledge.
\end{abstract}

\section{Introduction}\label{1}

The first direct observation of a gravitational wave (GW) event, GW150914, by the Advanced LIGO detectors~\cite{2015CQGra..32g4001L} in 2015, opened a new window for humanity to observe the universe~\cite{PhysRevLett.116.061102,PhysRevLett.116.241102}.
Since then, the LIGO and Virgo collaborations have reported the detection of over 90 GW events from compact binary coalescences (CBC) during three observing runs O1, O2, and O3~\cite{2019PhRvX...9c1040A,2021PhRvX..11b1053A,2024PhRvD.109b2001A,2023PhRvX..13d1039A}.
These events included binary back hole (BBH) mergers, binary neutron star (BNS) mergers, and neutron star black hole (NSBH) mergers. Notably, GW events involving BNS and NSBH systems are often accompanied by electromagnetic counterparts.  
For example, the BNS merger event GW170817, observed in 2017, was associated with multiple electromagnetic signals, marking the beginning of the era of multi-messenger astronomy~\cite{2017PhRvL.119p1101A,2017ApJ...848L..12A,2017ApJ...848L..15S}.

Currently, matched filtering remains one of the most widely used techniques for GW detection and has played a critical role in identifying signals from CBC sources~\cite{Allen:2005fk}. This method computes the signal-to-noise ratio (SNR) by matching the detector strain with a template bank.
The LIGO-Virgo-KAGRA (LVK) collaboration~\cite{2015CQGra..32b4001A,2019NatAs...3...35K} utilizes several well-established search pipelines, many of which are based on matched filtering. These include PyCBC~\cite{2021ApJ...923..254D}, GstLAL~\cite{PhysRevD.109.042008}, MBTA~\cite{2025CQGra..42j5009A}, and SPIIR~\cite{PhysRevD.105.024023}, which have been instrumental in CBC detection. In addition, there are pipelines like cWB~\cite{PhysRevD.93.042004} that detect unmodeled GW transients without relying on matched filtering.
However, as the template bank expands to cover a broader parameter space, the computational cost of matched filtering increases significantly~\cite{2024PhRvD.109d2005S,Harry:2016ijz}.
With the anticipated improvement in detector sensitivity and the expansion of the global detector network, hundreds of GW sources are expected to be observed annually~\cite{2020LRR....23....3A}.
These sources must be localized quickly to enable timely electromagnetic follow-up observations and maximize the potential for multi-messenger discoveries.
In addition to computational challenges, non-Gaussian transient noise, commonly referred to as glitches, frequently appears in the data~\cite{2016CQGra..33m4001A}. These glitches can be misidentified as GW signals, significantly impacting the sensitivity of GW searches~\cite{2019CQGra..36o5010C}.
To address these challenges, deep learning techniques have emerged as promising alternatives to traditional matched filtering. Among them, convolutional neural networks (CNNs) have demonstrated substantial potential~\cite{2023arXiv231115585Z}.

CNNs are deep learning algorithms inspired by the structure of the biological visual cortex. They were initially popularized in handwritten digit recognition~\cite{726791}. 
The core component of CNNs, the convolutional layer, employs learnable filters to extract hierarchical features through convolution operations with input data~\cite{2014arXiv1406.4773S,10.5555/2999134.2999257,2020arXiv200913120W,2023NucTe.209.1365R}. 
Interestingly, some researchers have pointed out that matched filtering can be understood as a specialized type of CNN that performs convolution operations using predefined templates~\cite{PhysRevD.105.043006,PhysRevD.101.104003,2022PhRvD.105h3013M}.
CNNs have attracted considerable attention in GW data analysis research, due to their powerful feature extraction capabilities. Although they are not yet commonly used in operational CBC detection pipelines, they have shown strong potential in experimental studies.
The early studies had proved the effectiveness of CNNs for GW searches, indicating that CNNs can classify the signal of BBH mergers from white noise and the performance is close to that of the matched filtering~\cite{2018PhRvD..97d4039G,2018PhRvL.120n1103G}.
Subsequent research has explored adaptations of these methods to more complex physical and observational scenarios~\cite{2020PhLB..80335330K,2019SCPMA..6269512F}.
Other studies have focused on optimizing training strategies and network architectures to further enhance model performance~\cite{Wang:2019zaj,2021PhLB..81236029W,2021Senso..21.3174M}.
Additionally, CNN-based approaches have been successfully extended to detect signals from a broader range of astrophysical sources, including BNS mergers~\cite{2020PhRvD.102f3015S,2021PhLB..81536161K,2021PhLB..81636185W,2021PhRvD.103j2003B}, NSBH mergers~\cite{2021PhRvD.104f2004Y,2023PhLB..84037850Q}, and eccentric CBC~\cite{2021ApJ...919...82W}.

Nevertheless, the full potential of CNNs in GW detection remains underexplored. Although CNNs can automatically extract high-level features without manual engineering~\cite{2017IGRSM...5d...8Z}, the physical interpretability of these learned features has received limited attention. Recent work has begun to address this gap, for instance, by applying dimensionality reduction techniques such as 
t-distributed stochastic neighbor embedding (t-SNE)  to visualize the final convolutional feature maps in two dimensions~\cite{2024PhRvD.109d3011S}. These visualizations revealed clear clustering between BBH signals and noise, suggesting that CNNs can inherently learn physically meaningful representations. However, the explicit characterization and utilization of such features remain open research questions. To further improve interpretability and performance, we propose a hybrid architecture that combines a CNN-based feature extractor with a random forest (RF) classifier, which we refer to as CNN-RF. Unlike prior works that directly connect a classifier (such as RF) to the final convolutional or fully connected layer~\cite{kwak2021potential,Sukanya2024HybridCA,ijgi10040242,rs15030728}, our method further extracts four interpretable handcrafted features from the final convolutional layer. By combining these features with the CNN output probability, the RF classifier can establish more effective decision boundaries. This dual-strategy approach not only enhances the robustness and interpretability of the model but also provides deeper insights into the underlying physical phenomena.
Moreover, this approach can be rapidly extended to other related studies, due to the modularity design. In recent years, numerous works have introduced advanced CNN-based models, such as Residual Networks (ResNet)~\cite{2021PhLB..81236029W}, demonstrating their effectiveness in GW detection. Using these advanced models, which can extract more discriminative features~\cite{SIMON20201680}, the proposed hybrid framework has the potential to significantly improve signal discrimination capabilities.

This paper is structured as follows. In Section~\ref{2}, we describe the datasets used in our work. Section~\ref{3} introduces the hybrid CNN-RF model architecture and outlines its training process. A detailed discussion of the evaluation methodology is given in Section~\ref{4}. We present and analyze the experimental results in Section~\ref{5}. Finally, Section~\ref{6} summarizes the key findings and concludes the paper.

\section{Data Preparation}\label{2}

In this work, three datasets were generated. The first dataset contains GW signals and noise. The second includes GW signals, noise, and glitches. The third one is the long-duration data consisting of one week of data with randomly injected GW signals.
Our focus is solely on the GW signals from the BBH mergers.

\subsection{Dataset 1}\label{2.1}

The dataset 1 was generated by the open-source project \textbf{ggwd} by Gebhard et al.~\cite{2019PhRvD.100f3015G}, which can generate synthetic GW signals with a real LIGO data. This dataset contains two classes data, simulated BBH GW signals and pure noise, all the data with different GPS time. The sampling rate of the dataset is 4096~Hz and the duration is 8 seconds. All the simulated GW signals include the inspiral, merger, and ringdown phases, with the merger occurring at 5.5 s.
The synthetic signal can be described simply as
\begin{equation}
s(t) =  n(t) + \alpha h(t),
\label{eq:1}
\end{equation}
where $n(t)$ is the background noise, $h(t)$ is the simulated GW waveform, and $\alpha$ is the scale factor that decides the injection SNR of this synthetic signal. The background noise is the first observing run (O1) data of LIGO excluding known GW events, and has passed standard data quality checks, ensuring its suitability for GW analysis~\cite{2016CQGra..33m4001A}. 
The GW waveforms are simulated by the effective-one-body model \textbf{SEOBNRv4}~\cite{Bohe:2016gbL}, and the parameters are randomly sampled in the given interval. The masses of black holes are in the range of $(5M_{\odot },80M_{\odot })$, the $z$-components of the spin of the merging black holes in $(0,0.998)$, the polarization angles in $(0,2\pi)$. The values of coalescence phase and inclination angle are sampled jointly from a uniform distribution over a sphere. The values of the right ascension and the declination are sampled together from a uniform distribution over the sky. 
The distance between the detectors and the source is fixed at 100 Mpc. The injection SNR is randomly sampled from the range $(5,20)$ to rescale the waveform by the scale factor, indicating that the source appears to move closer or farther from the detectors~\cite{2019PhRvD.100f3015G}.

Every sample of dataset contains two strains representing LIGO's Hanford (H1) and Livingston (L1) detectors, respectively. 
Based on the sky location of each injection, the appropriate inter-detector time delay was calculated and applied during the waveform injection process to maintain physical consistency across the two detectors.
The dataset was whitened using \textbf{PyCBC v2.5.1}, where the power spectral density (PSD) was estimated from the same segment of O1 noise data via Welch's method, without applying the normalization factor~\cite{2024zndo..10473621N}. Each PSD estimate used a segment duration of 4 seconds (with 50\% overlap) and a Hann window. After whitening, a high-pass filter at 20~Hz was applied to eliminate the influence of low-frequency Newtonian noise~\cite{2016CQGra..33u5004U}. 
We generated 108800 samples, and one-half of samples are the simulated GW signals, the other samples are the pure noise.

\subsection{Dataset 2}\label{2.2}
This dataset includes not only GW signals and noise, but also glitches. The process of generating GW signals and noise is the same as that used for the previous dataset. The process of generating glitches involves choosing the peak time of the glitches and selecting the corresponding strain data. The data were then whitened, high-passed, and cut in the same way as for Dataset 1.  
The glitches were taken from the first observing run (O1) only, to ensure that all three classes share the same background noise characteristics.
The peak times of the glitches were obtained from GravitySpy, an open-source program dedicated to LIGO glitch classification, and the corresponding glitch segments were selected from the O1 subset of the catalog~\cite{2023CQGra..40f5004G}.
We generated a total of 60,000 samples, with 20,000 samples for each type of data.

\subsection{Dataset 3}\label{2.3}

Unlike the previous two datasets, this one consists of long-duration data. We selected one week of data from the O1 run to serve as the background dataset. The one week of data contains multiple non-contiguous data segments (each 4096 seconds in duration), with a total duration of one week. These segments were specifically chosen to exclude known GW events and contain simultaneous strain recordings from both the H1 and L1 detectors.
We then randomly inserted 40,000 simulated GW signals. To prevent signal overlap, we enforced a minimum time separation of 10 seconds between successive injections. 
The dataset processing methods remain consistent with those used previously. We refer to the original one-week dataset as the `background set' and the dataset with inserted signals as the `foreground set.'

\section{Hybrid Model Architecture}\label{3}

\begin{figure}[htpb]
    \centering
    \includegraphics[width=0.6\linewidth]{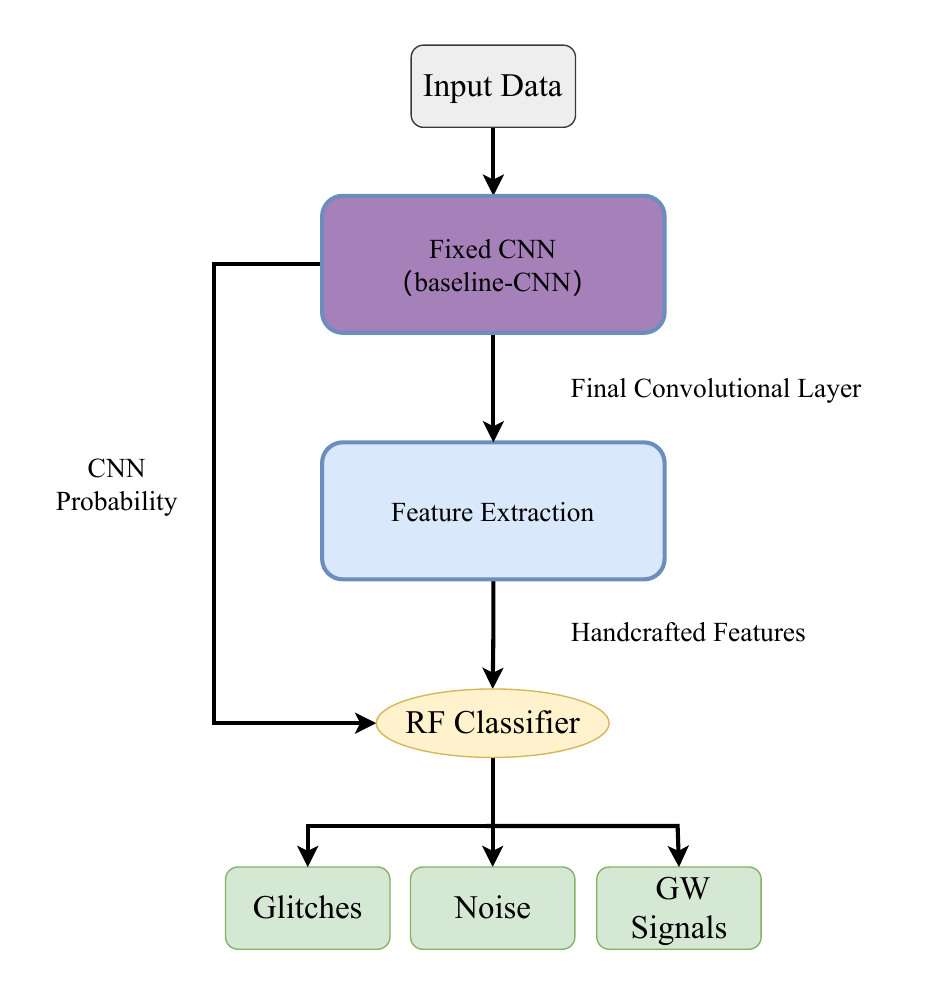}
    \caption{Architecture of the CNN-RF model. The raw input data is first processed by a fixed CNN. Features are then extracted from the output of its final convolutional layer, from which four handcrafted features are computed. These features, along with the $\mathbf{CNN\ probability}$, are concatenated and fed into an RF classifier for the final classification decision.}
    \label{FIG.1}
\end{figure}

The hybrid CNN-RF model architecture is illustrated in Figure~\ref{FIG.1}.
In this model, a fixed CNN is used to guide feature extraction, while a RF is employed for classification. The CNN is first trained to distinguish between GW signals and noise. This trained network, referred to as the baseline-CNN, is then fixed. From its final convolutional layer, four handcrafted features are extracted. Additionally, the CNN output probability—representing its confidence that a given input contains a GW signal (hereafter referred to as the $\mathbf{CNN\ probability}$)—is included. These five features are then used as inputs to train the RF classifier to differentiate among GW signals, noise, and glitches.

\subsection{CNN Module}\label{3.1}
                       
CNN is a specific category of deep learning algorithms, particularly adept at processing grid-structured data such as images and temporal signals. It is capable of capturing useful feature representations~\cite{Lecun1998}. As the depth of convolutional layers increases, CNN can capture more abstract and complex features in deeper layers~\cite{lecun2015deep}. The architecture of a CNN primarily comprises convolutional layers, pooling layers, and fully connected layers. Convolutional layers are the cornerstone of CNN, designed mainly for feature extraction from input data. These layers apply a set of learnable filters or convolution kernels to scan the input, where each filter performs dot multiplication with local regions of the input to produce feature maps. Following these feature maps, a non-linear activation function, such as the Rectified Linear Unit (ReLU), is applied to introduce non-linearity into the model, enhancing its expressive power. Pooling layers, typically placed after convolutional layers, reduce the spatial dimensions of the feature maps by selecting maximum (max pooling) or average (average pooling) values over local neighborhoods. This process not only decreases the number of parameters and computational load for subsequent layers but also improves the model's robustness against variations in the input. Fully connected layers are situated towards the end of the CNN architecture, tasked with integrating all features extracted by preceding layers to perform the final classification.

Building upon these principles, our baseline-CNN architecture (Table~\ref{tab:architecture}) consists of seven convolutional layers, three pooling layers, and one fully connected layer. Each convolutional layer is followed by batch normalization and ReLU activation function. All convolutions are implemented as 2D convolutional layers with kernel sizes of the form $(1 \times k)$, where $k$ varies across layers. The first dimension of the kernel corresponds to the detector axis and is fixed to 1, ensuring that convolution is applied independently to each detector's time series without cross-detector interaction. No padding is used in any layer.
The input tensor has the shape of $(batch\_size,1,2,4096)$, where the second dimension serves as a singleton placeholder for framework compatibility, the third dimension corresponds to the H1 and L1 detectors, and the fourth dimension contains 4,096 temporal samples per channel, corresponding to 1-second duration signals at a sampling rate of 4096~Hz.

\begin{table}[htbp]
\centering
\captionsetup{justification=centering} 
\caption{The Baseline-CNN Architecture.}
\resizebox{0.6\textwidth}{!}{ 
\begin{tabular}{ccccc}
\toprule
& Layer & Kernel Size & Stride  & Shape \\
\midrule
& Input & & & $1 \times 2 \times 4096$ \\
\midrule
1 & Convolution & (1,32) & (1,1)  & $8 \times 2 \times 4065$ \\
2 & Max pooling & (1,8)  & (1,8)  & $8 \times 2 \times 508$ \\
3 & Convolution & (1,16) & (1,1)  & $16 \times 2 \times 493$ \\
4 & Convolution & (1,16) & (1,1)  & $16 \times 2 \times 478$ \\
5 & Convolution & (1,8) & (1,1)  & $32 \times 2 \times 471$ \\
6 & Max pooling & (1,6) & (1,6)  & $32 \times 2 \times 78$ \\
7 & Convolution & (1,8) & (1,1)  & $64 \times 2 \times 71$ \\
8 & Convolution & (1,4) & (1,1)  & $128 \times 2 \times 68$ \\
9 & Convolution & (1,4) & (1,1)  & $128 \times 2 \times 65$ \\
10 & Max pooling & (1,4) & (1,4)  & $128 \times 2 \times 16$ \\
11 & Flatten & - & -  & 4096 \\
12 & Fully connected & - & -  & 2 \\
\midrule
& Output (Softmax) & & & 2 \\
\bottomrule
\end{tabular}
}
\label{tab:architecture}
\end{table}

\subsection{RF Classifier}\label{3.2}

Most hybrid models combine CNN with SVM (Support Vector Machine) or CNN with RF for classification tasks. In our work, we chose RF as the classifier. The most important reason for this choice is that RF can provide feature importance, which indicates the contribution of each feature in the decision-making process. This allows us to effectively evaluate the efficiency of our handcrafted features.

The RF is an ensemble learning algorithm that consists of multiple decision trees. Decision trees are interpretable machine learning models that use a tree-based architecture to recursively split data by optimizing certain criteria, such as Gini impurity, for classification tasks. 
Gini impurity is a measure of the purity of a set, defined as the probability of misclassifying a randomly selected element if it were labeled according to the distribution of classes in the set~\cite{loh2011classification}. A lower Gini impurity indicates a higher degree of set purity.
However, as the depth of the decision tree increases, the decision boundary becomes more complex, making the model more prone to overfitting on the training data. RF mitigates this issue by constructing an ensemble of decision trees through a process called bagging. Each tree is trained on a random subset of the data sampled with replacement (bootstrap sampling), and predictions are made by aggregating the outputs of individual trees—using majority voting for classification tasks or averaging for regression tasks—thereby improving generalization and reducing the risk of overfitting~\cite{2001MachL..45....5B}.
A significant advantage of RF is its inherent capability to measure feature importance using permutation-based methods. Features with higher importance scores have a greater impact on decision boundaries within the ensemble.

We employed the \textbf{RandomForestClassifier} from the \href{https://scikit-learn.org}{scikit-learn} library to build our RF model. The model was configured with 1000 decision trees, the criterion set to Gini impurity, the max depth of 10 and a fixed random state of 1. All other hyperparameters were set to their default values as provided by scikit-learn.

\subsection{Handcrafted Features Extraction}\label{3.3}

CNNs are effective feature extractors, capable of capturing high-level features through their deep layers. Research indicates that higher layers tend to generate more discriminative features~\cite{zeiler2014visualizing}. Despite this capability, our understanding of the features extracted by these networks, particularly in the deeper layers, remains limited. In image processing, CNNs typically extract low-level features such as edges, textures, and colors in their shallow layers, while deeper layers capture more abstract, semantic-level information~\cite{krizhevsky2017imagenet}.
Motivated by this, we aim not only to leverage the CNN's extracted features but also to interpret them more effectively. To achieve this, we extract handcrafted features from the final convolutional layer, enabling enhanced classification and interpretability.

From the feature maps of the final convolutional layer, we extracted four handcrafted features: $\mathbf{SNR}$, $\mathbf{waveform\ overlap}$, $\mathbf{variance}$, and $\mathbf{peak\ amplitude}$. Among them, $\mathbf{variance}$ and $\mathbf{peak\ amplitude}$ are common handcrafted features that can represent the statistical distribution of data. The $\mathbf{variance}$ measures the spread or dispersion of the data points around the mean, giving an indication of how much the values deviate from their average. The $\mathbf{peak\ amplitude}$ identifies the maximum value within the dataset.
The inspiration for $\mathbf{SNR}$ and $\mathbf{waveform\ overlap}$ comes from matched filtering technique. 
GW signals recorded by detectors at different locations share similar characteristics due to their common astrophysical origin and coherent propagation. Thus, the CNN-extracted features from such data should also exhibit high similarity across detectors. 
Based on this assumption, we quantify the cross-detector similarity using the $\mathbf{SNR}$ and $\mathbf{waveform\ overlap}$ metrics.

From Table~\ref{tab:architecture}, the output of the final convolutional layer has a shape of (128, 2, 65), representing 128 feature maps, each of size $2\times65$. For clarity, we define the dimension of size 65 corresponding to the H1 detector as the H1-feature, and the other dimension (also size 65) as the L1-feature, referring to the L1 detector. In the $i$-th feature map, the H1-feature is denoted by $h_i$, and L1-feature by $l_i$.

The $\mathbf{SNR}$ is defined as the average over all feature maps of a matched-filter-like statistic, which captures the  coherent amplitude consistency between the H1- and L1-features.
For each feature map $i$ (from $1$ to $N=128$), we compute two inner products in the frequency domain
\begin{align}
\left \langle \tilde{h}_{i} \mid \tilde{l}_{i}^{*} \right \rangle  = 4 \Re\left\{\int_{0}^{\infty} \tilde{h}_{i}(f)\tilde{l}_{i}^{*}(f) df \right\}, \quad
\left \langle \tilde{l}_{i} \mid \tilde{h}_{i}^{*} \right \rangle  = 4 \Re\left\{\int_{0}^{\infty} \tilde{l}_{i}(f)\tilde{h}_{i}^{*}(f) df \right\},
\end{align}
where $\tilde{h}_{i}(f)$ and $\tilde{l}_{i}(f)$ are the Fourier transforms of the H1- and L1-features from the $i$-th feature map. The symbol $^*$ denotes complex conjugation.
For each feature map, we define the single-detector-normalized statistic $\mathbf{SNR}_{i}$ as the maximum of the normalized inner products in both directions
\begin{align} 
\mathbf{SNR}_{i} &= \max\left( \frac{ \left | \left \langle \tilde{h}_{i} \mid \tilde{l}_{i}^{*} \right \rangle \right | }{ \sqrt{\left \langle \tilde{h}_{i} \mid \tilde{h}_{i}^{*} \right \rangle}} , 
\frac{ \left | \left \langle \tilde{l}_{i} \mid \tilde{h}_{i}^{*} \right \rangle \right | }{ \sqrt{\left \langle \tilde{l_{i}} \mid \tilde{l}_{i}^{*} \right \rangle} }\right). 
\end{align}
The final $\mathbf{SNR}$ is obtained by averaging over all $N$ feature maps
\begin{align} 
\mathbf{SNR} &= \frac{1}{N} \sum_{i=1}^{N}\mathbf{SNR}_{i}.
\end{align}
Unlike traditional SNR calculations, we do not divide by PSD, as the input data are whitened beforehand. Furthermore, to account for detector-specific differences in signal amplitude, we compute the inner product in both directions and take the maximum value.

The $\mathbf{waveform\ overlap}$ is a normalized inner product that measures the phase similarity between the H1- and L1-features. For each feature map $i$, the $\mathbf{waveform\ overlap}_i$\footnote{The $\mathbf{SNR}_{i}$ and $\mathbf{waveform\ overlap}_{i}$ were computed using the \texttt{matched\_filter} and \texttt{overlap} functions in \texttt{PyCBC}, available in the \href{https://pycbc.org/pycbc/latest/html/pycbc.filter.html}{\texttt{pycbc.filter}} module.} is defined as
\begin{align}
\mathbf{waveform\ overlap}_{i} &= \frac{ \left | \left \langle \tilde{h}_{i} \mid \tilde{l}_{i}^{*} \right \rangle \right | }{ \sqrt{\left \langle \tilde{h}_{i} \mid \tilde{h}_{i}^{*} \right \rangle} .\sqrt{\left \langle \tilde{l}_{i} \mid \tilde{l}_{i}^{*} \right \rangle} }. 
\end{align}
The final $\mathbf{waveform\ overlap}$ is obtained by averaging across all feature maps
\begin{align}
\mathbf{waveform\ overlap} = \frac{1}{N} \sum_{i=1}^{N}\mathbf{waveform\ overlap}_{i}.
\end{align}
This metric is bounded between 0 and 1, where a value close to 1 indicates a higher degree of phase alignment between the two detectors. Compared to $\mathbf{SNR}_{i}$, which is normalized with respect to only one detector at a time (either H1 or L1), the $\mathbf{waveform\ overlap}_{i}$ is symmetrically normalized with respect to both detectors. This symmetric normalization ensures that the resulting value is invariant under amplitude rescaling of either input, thereby isolating the phase similarity component. In contrast, $\mathbf{SNR}_{i}$ is designed to reflect coherent amplitude consistency and is therefore asymmetric by construction, taking the maximum from two one-sided normalizations. These complementary definitions enable the joint characterization of both amplitude and phase alignment between H1- and L1-feature representations.

The $\mathbf{variance}$ measures the normalized statistical difference between the H1- and L1-features. For each feature map $i$, we define the pointwise difference vector as
\begin{align}
x_j = h_j - l_j, \quad j = 1, \dots, M,
\end{align}
where $h_j$ and $l_j$ denote the $j$-th elements of the H1- and L1- feature vectors, respectively, and $M$ is the feature vector length (in our case, $M = 65$). To remove scale dependence, the difference vector is min-max normalized
\begin{align}
\hat{x}_j = \frac{x_j - \min(x)}{\max(x) - \min(x)}.
\end{align}
We then compute the $\mathbf{variance}_{i}$ of the normalized differences for each feature map
\begin{align}
\mathbf{variance}_i = \frac{1}{M} \sum_{j=1}^{M} (\hat{x}_j - \mu)^2, \quad \mu = \frac{1}{M} \sum_{j=1}^{M} \hat{x}_j.
\end{align}
The final $\mathbf{variance}$ is obtained by summing across all feature maps
\begin{align}
\mathbf{variance} = \sum_{i=1}^{N} \mathbf{variance}_i.
\end{align}
A lower $\mathbf{variance}$ value indicates stronger similarity between the H1- and L1-features, suggesting coherent signal presence, while a higher value may reflect non-Gaussian noise or detector-specific artifacts.

The $\mathbf{peak\ amplitude}$ captures the maximum absolute deviation between the H1- and L1-features and reflects pronounced detector-specific differences in signal strength. For each feature map, we compute 
\begin{align}
\mathbf{peak\ amplitude}_i = \sum_{i=1}^{M} \max(\left | x_{j} \right | ).
\end{align}
The final $\mathbf{peak\ amplitude}$ is defined as the average over all feature maps
\begin{align}
\mathbf{peak\ amplitude} = \frac{1}{N} \sum_{i=1}^{N} \mathbf{peak\ amplitude}_i.
\end{align}
While $\mathbf{variance}$ captures the overall statistical fluctuation after normalization, $\mathbf{peak\ amplitude}$ retains absolute scale information, making it a complementary feature for identifying localized, high-amplitude discrepancies between detectors.

\subsection{Training Process}\label{3.4}

We first trained the baseline-CNN on Dataset 1, of which  $80\%$ is used for the training, $10\%$ for the validation, $10\%$ for the testing.
Each original data sample consists of an 8-second duration, which needs to be segmented into a 1-second window to match the input dimension of the CNN. 
We randomly segmented each sample into a 1-second window, ensuring that the merger time of the GW signal falls within the interval $(\frac{1}{3}, \frac{2}{3})$s, to prevent the model from overly relying on specific times and locations.
Notably, this segmentation is performed only once during the initial training phase, and subsequent training iterations utilize the pre-segmented samples.
We used the cross-entropy loss as the loss function, and optimized the network using the Adam optimizer~\cite{Kingma2014AdamAM} with an initial learning rate of 0.001. 
A cosine annealing learning rate scheduler~\cite{Loshchilov2016SGDRSG} was applied during training, with the period parameter set to $T_{\text{max}} = 20$ epochs, allowing the learning rate to gradually decay following a cosine schedule. 
The baseline-CNN was trained for 20 epochs with a batch size of 64, and the final model was selected based on the lowest training loss. 
All experiments were conducted in a Docker environment using a \textbf{NVIDIA 3060Ti GPU} with \textbf{CUDA 12.3.2} and \textbf{PyTorch 2.2.1}, requiring approximately 80 hours of computation time.

Before training the RF classifier, the pre-trained baseline-CNN served as the feature extractor, and extracted four handcrafted features, following the procedure described in Section~\ref{3.3}.
This process generated four handcrafted features from three types of data: noise, glitches, and GW signals from Dataset 2. These features (four handcrafted features plus $\mathbf{CNN\ probability}$) were then used as input to train the RF classifier. The dataset was then split into $80\%$ training, and $20\%$ testing sets. The RF classifier architecture (Section~\ref{3.2}) was implemented with \textbf{scikit-learn 1.4.1} in \textbf{Python 3.9.19}. The experiments were conducted on a machine with an \textbf{Intel(R) Core(TM) i5-12400F CPU @ 2.50GHz} and \textbf{16 GB RAM}.

\section{Evaluation}\label{4}
\subsection{Sensitivity  Metrics}\label{4.1}

We evaluated the models using Dataset 3, which contains long-duration time-series data.
In contrast to the 1-second segments used in training, real detector data are significantly longer. Assessing model performance on such data provides a more realistic indication of effectiveness in practical GW detection scenarios. All metrics were computed using one week of data.

We employ two key metrics: the sensitive fraction and the sensitive distribution.
The sensitive fraction is defined as a function of the false-alarm rate (FAR), representing the proportion of correctly identified GW signals under a given false positive constraint. This metric is widely adopted in GW searches~\cite{2016CQGra..33u5004U} and is related to the sensitive distance metric~\cite{2023PhRvD.107b3021S}. 
However, we did not compute the sensitive distance in this study, as the distance distribution of the injected signals is unclear. Although the initial distance between the source and the detector is fixed at 100 Mpc during signal injection, we vary the signal strength through the scaling factor $\alpha$ based on the injection SNR, which mimics the effect of the source moving closer to or farther away from the detector. However, since injection SNR alone is not sufficient to uniquely determine the actual source distance, this approach does not allow us to precisely infer the true distance.
The sensitive distribution characterizes the sensitivity as a function of the injection SNR at a fixed FAR. It illustrates how detection performance varies with the strength of the injected GW signals, providing insight into model capability across different signal regimes.

Dataset 3 consists of both background and foreground sets. When the search algorithm processes the dataset, it returns a list of events. Each event contains a ranking confidence, a GPS time, and a timing accuracy. The detailed process is described in Section~\ref{4.2}. The events in the background set are called background events, and those in the foreground set are called foreground events. 
To calculate the relationship between sensitivity and FAR, it is necessary first to determine FAR and sensitivity as functions of the ranking confidence. Then, both are evaluated simultaneously under the same ranking confidence value and the results are combined. By using ranking confidence, of all background events for evaluation, it is ensured that each point is a unique false positive event.

The FAR is calculated solely from background events, all of which are confirmed false positives. 
The form of FAR $\mathcal{F}$ is
\begin{equation}
\mathcal{F}=\frac{N_{\mathrm{FP}, \mathcal{R}}}{T},
\end{equation}
where the $N_{\mathrm{FP}, \mathcal{R}}$ is the number of background events with ranking confidence greater than or equal to $\mathcal{R}$, and $T$ is the total duration of the dataset. 
Due to memory constraints and to reduce computational costs, in this study we generated only one week of data. However, the FAR is typically reported for a one-month period in practical scenarios. Assuming that the background noise is stationary, the monthly FAR can be extrapolated from the weekly result using $\mathcal{F}{\mathrm{month}} = \mathcal{F}{\mathrm{week}} \times (T_{\mathrm{month}} / T_{\mathrm{week}})$, where $ T_{\mathrm{month}} $ and $ T_{\mathrm{week}} $ denote the durations of one month and one week, respectively. 

The sensitive fraction of FAR $\mathcal{F}$ is
\begin{equation}
S(\mathcal{F}) = \frac{N_{\mathrm{I},\mathcal{F}}}{N_{\mathrm{I}}},
\end{equation}
where $N_{\mathrm{I},\mathcal{F}}$ is the number of detected injected signals at a FAR $\mathcal{F}$, $N_{\mathrm{I}}$ is the total number of injected signals. The FAR $\mathcal{F}$ corresponds to a ranking confidence of background events.
An injected signal is considered detected if a returned event’s ranking confidence exceeds the threshold corresponding to $\mathcal{F}$, and its GPS time falls within a predefined timing accuracy around the injection time.

\subsection{Search Process}\label{4.2}

Detecting signals from long-duration data poses a significant challenge for the model, as it must handle more complex situations. This complexity arises from the increased likelihood of encountering various types of noise, glitches, making accurate detection and classification more difficult. To address these challenges, we designed an adaptive trigger-based detection framework that combines a sliding window approach with cluster-based post-processing.

The search begins with a sliding window operation that extracts fixed-length (1 second) segments from the time-series data. This sliding window approach processes time-series data by extracting fixed-duration segments (window size = 1 second) and inputting them sequentially into the model. After each calculation, the window advances by a predefined moving step ($\Delta t = 0.1$ seconds). When the model's output exceeds a predefined threshold (set to 0.3 in our experiments), the window is classified as a trigger, and the corresponding GPS time (taken as the window’s midpoint) and output value are recorded.

\begin{figure}[htpb]
    \centering
    \includegraphics[width=0.8\linewidth]{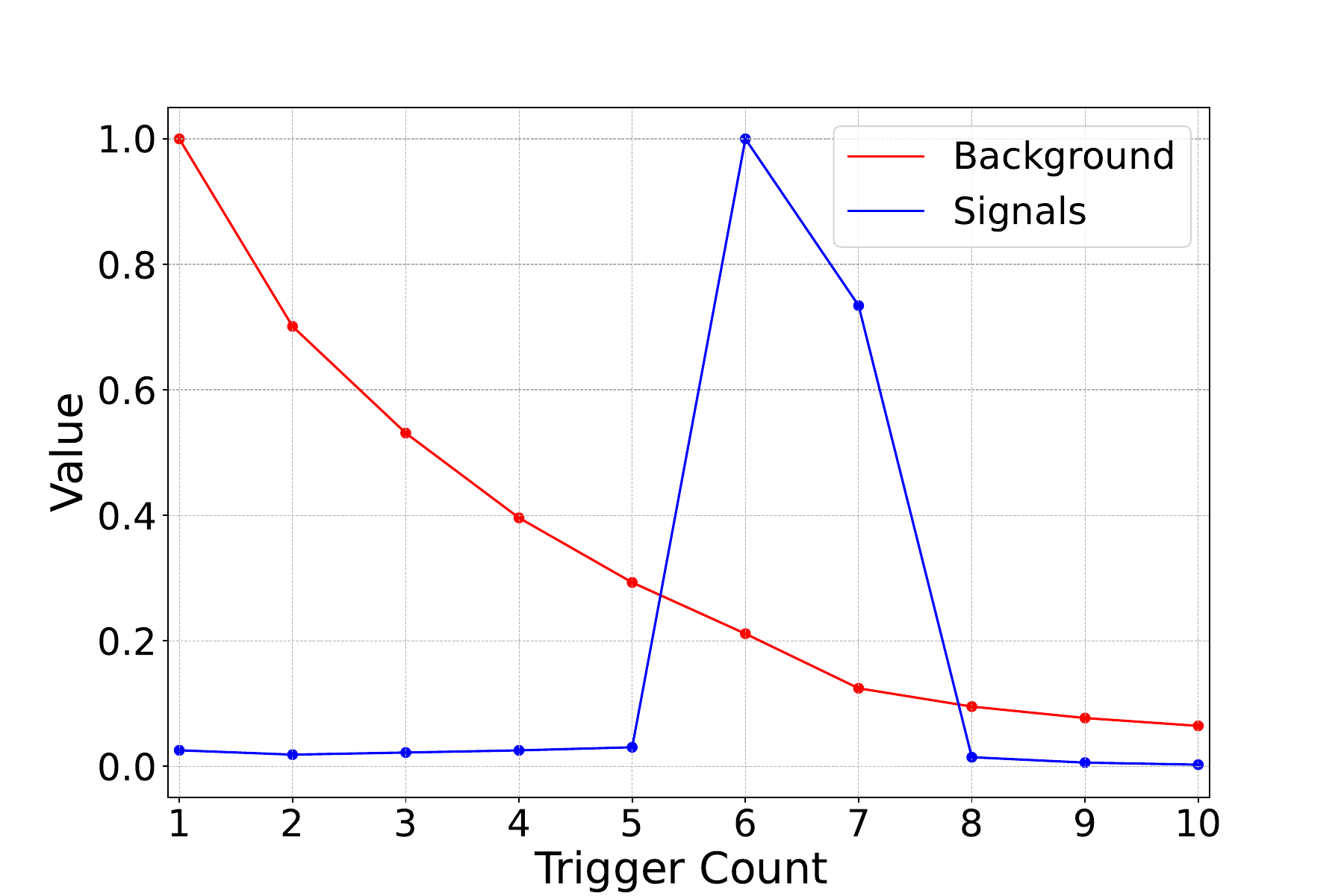}
    \caption{Trigger count distribution per cluster for the background and injected signals. All values are normalized using max-normalization. The red line corresponds to the background set, and the blue line represents the injected signals.}
    \label{FIG.2}
\end{figure}

Due to overlapping of sliding windows, the model may generate multiple consecutive triggers when scanning the same signal region. 
However, many of these triggers are caused by glitches, which differ from GW signals in terms of trigger frequency. Glitches typically have shorter durations than GW signals, leading to fewer overlapping windows detecting them, consequently fewer associated triggers.
To enhance the robustness of the search algorithm, we implemented a cluster-based post-processing method.
We cluster the triggers based on their GPS time, grouping triggers that occur consecutively into the a cluster.
If a cluster contains more than a threshold number of triggers, it is identified as a candidate GW event.
We empirically set this threshold to 6, based on the observed distributions of trigger counts for background and injected signals in Dataset 3. 
Figure~\ref{FIG.2} shows the trigger count distributions for both datasets.
The intersection point of the two curves lies between 5 and 6, making 6 a reasonable threshold that minimizes false alarms while preserving high sensitivity.

Each identified cluster corresponds to a potential GW signal detection. 
The maximum trigger value within a cluster indicates the confidence level of the detection. The GPS time of the event is defined as the GPS time of the trigger with the maximum value. This maximum value becomes the event's significance score, and its corresponding GPS time is recorded as the event’s occurrence time. 
To assess the temporal precision of our search algorithm, we define a timing accuracy parameter, $\Delta t$, which quantifies how closely the algorithm can localize the true signal time. The algorithm assigns a candidate GPS time $t_0$ to each detected signal, while the most closely matching injected signal has a known GPS time $t_\mathrm{I}$. If the absolute difference satisfies $\left| t_0 - t_\mathrm{I} \right| < \Delta t$, the injection is considered successfully detected. A smaller $\Delta t$ implies a stricter requirement on temporal accuracy. In this work, we set $\Delta t = 0.5$ seconds, corresponding to half of the model’s input window length, which provides a practical balance between localization precision and tolerance.

After clustering, we obtain a list of detected events. This list contains three one-dimensional datasets of equal size named `ranking confidence', `GPS time', and `timing accuracy'.
\begin{itemize}
\item \textbf{ranking confidence}: This dataset contains the values of the events, ranked from high to low. These values represent the confidence levels or significance of each event.
\item \textbf{GPS time}: This dataset contains the GPS times corresponding to the events in the `ranking confidence' dataset. 
\item \textbf{timing accuracy}: This dataset specifies the temporal precision for each event's detection. In this study, all values are set to 0.5 seconds.
\end{itemize}

\section{Results and Discussion}\label{5}

\subsection{Visualization}\label{5.1}

In this subsection, we analyze the features extracted by the baseline-CNN. To gain deeper insights into what the CNN learns, we visualized the feature maps produced at each convolutional layer. Feature visualization~\cite{erhan2009visualizing,olah2017feature,simonyan2013deep} is a powerful method to help people explain how CNN works. Specifically, our methodology involved capturing the feature maps generated by each layer in response to given inputs, thereby enabling an analysis of the features learned by each layer.
To achieve this, we defined a Hook class designed to intercept and record both input and output features during the forward pass through the network. By applying this technique, we were able to extract the output tensors from targeted convolutional layers and subsequently visualize these outputs. 

We focused on visualizing the input layer and the first two feature maps of each convolutional layer. To better understand what types of features are extracted, we selected four representative input samples: (1) a simulated BBH waveform ($m_1 = 30M_{\odot }$, $m_2=30M_{\odot }$); (2) a synthetic GW signal (waveform + noise, injection SNR =16); (3) a noise; (4) a glitch (event time=1132401286.3). Notably, the noise used in the synthetic GW signal is identical to the one used in the noise sample.

\begin{figure}[ht]
    \centering

    \leavevmode
    \raisebox{0pt}{\includegraphics[width=0.495\linewidth]{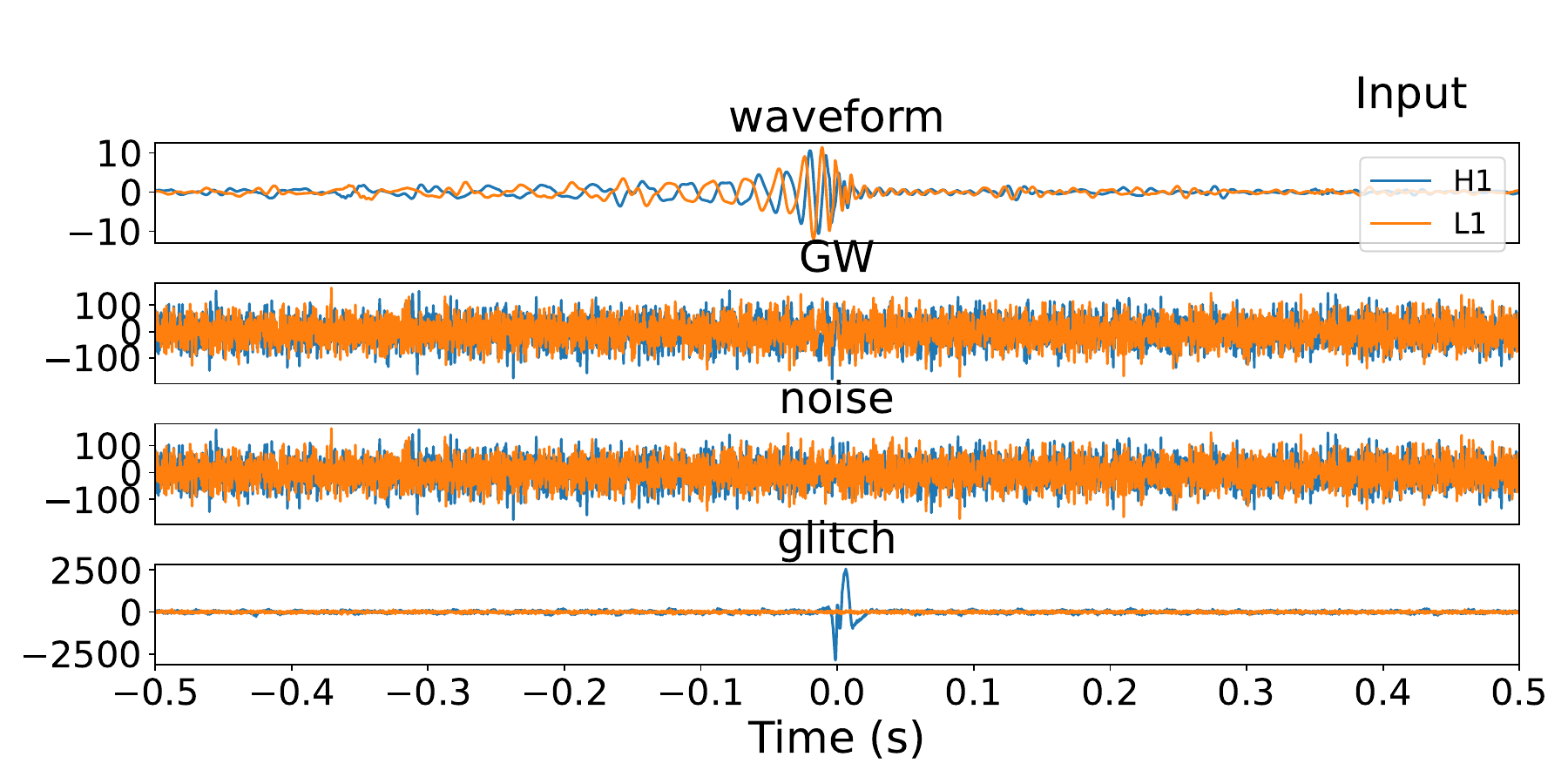}}%
    \hfill
    \raisebox{0pt}{\includegraphics[width=0.48\linewidth]{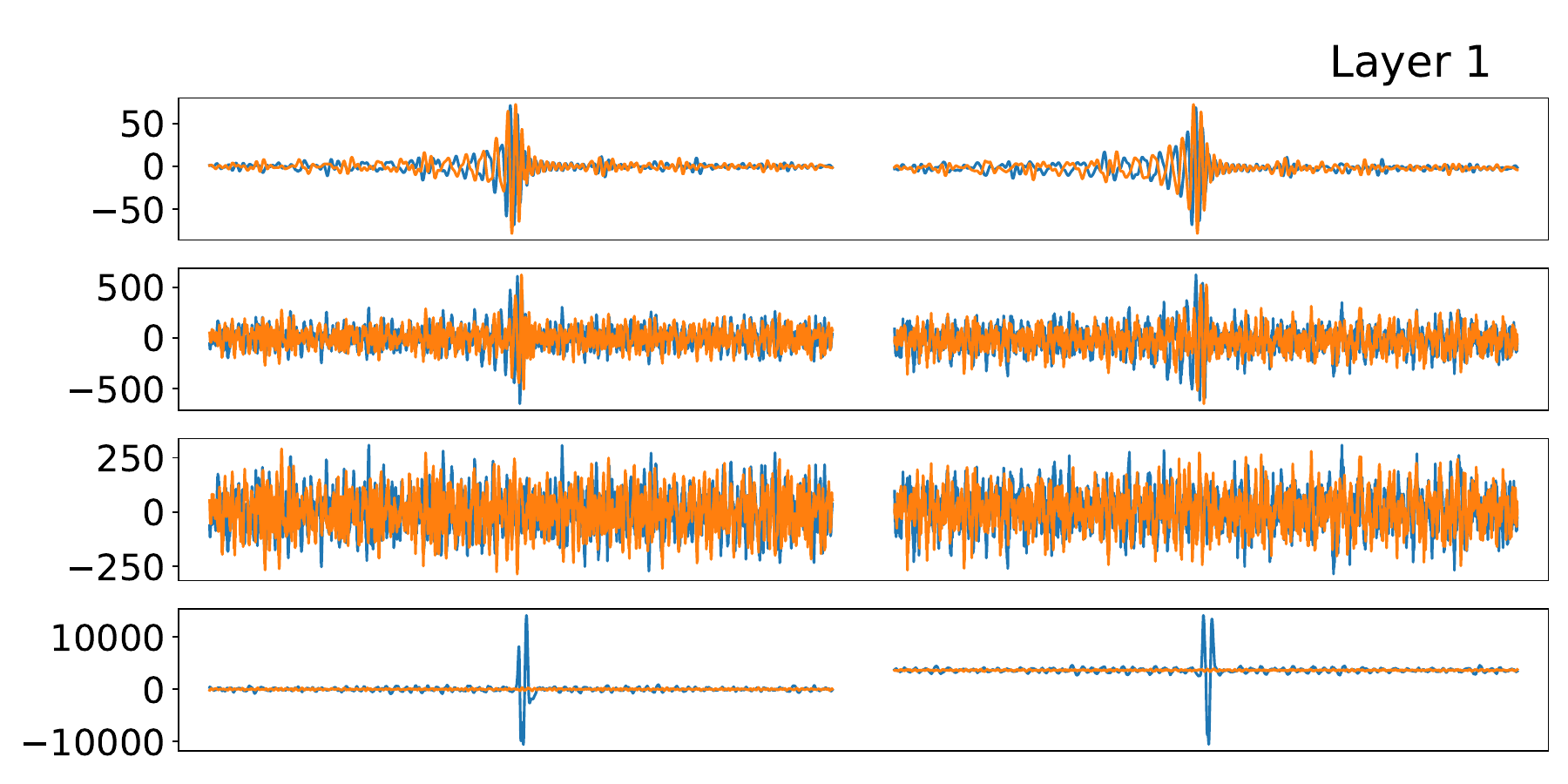}}%
    
    \vspace{1pt}\par

    \leavevmode
    \raisebox{0pt}{\includegraphics[width=0.48\linewidth]{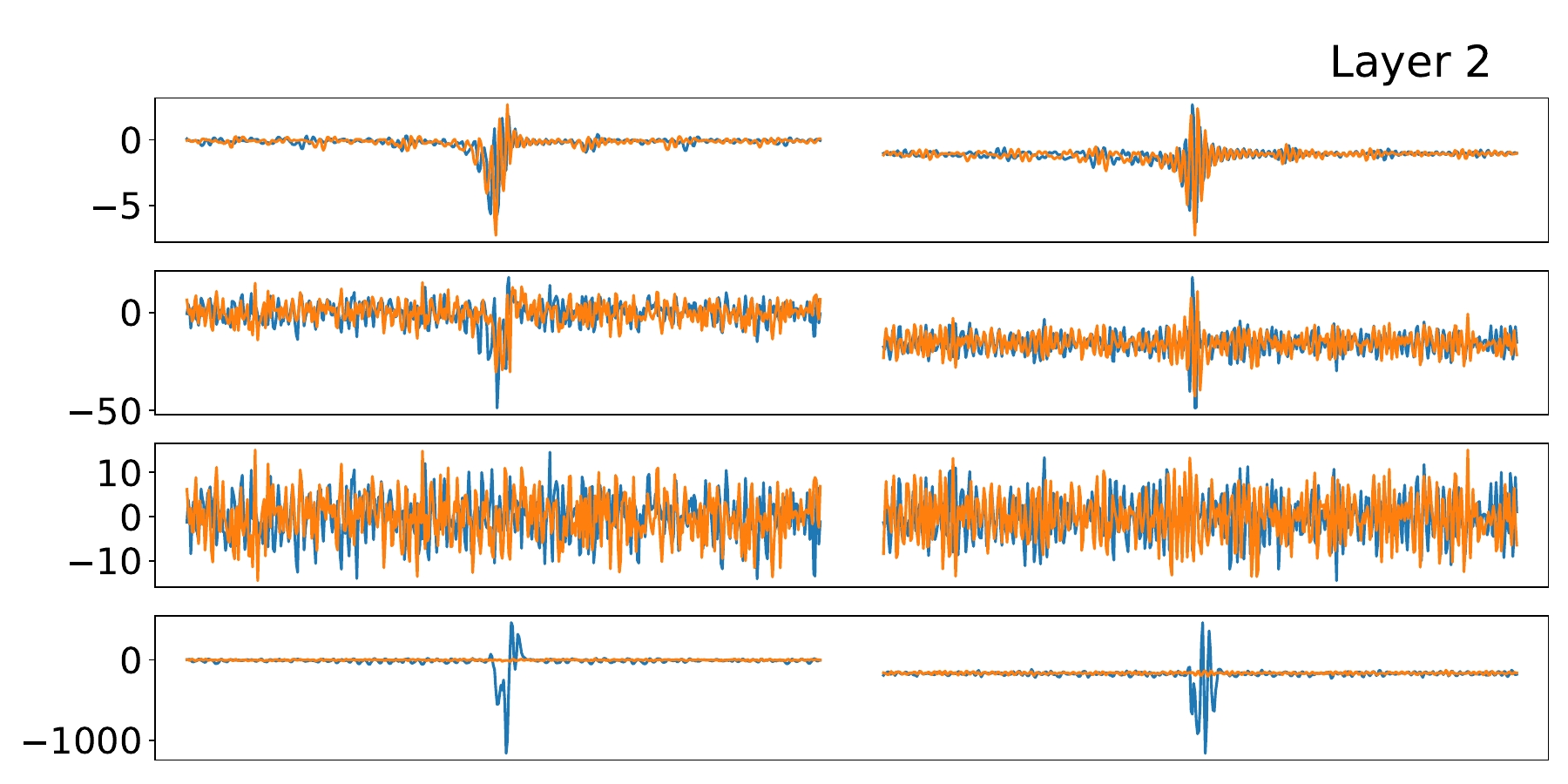}}%
    \hfill
    \raisebox{0pt}{\includegraphics[width=0.48\linewidth]{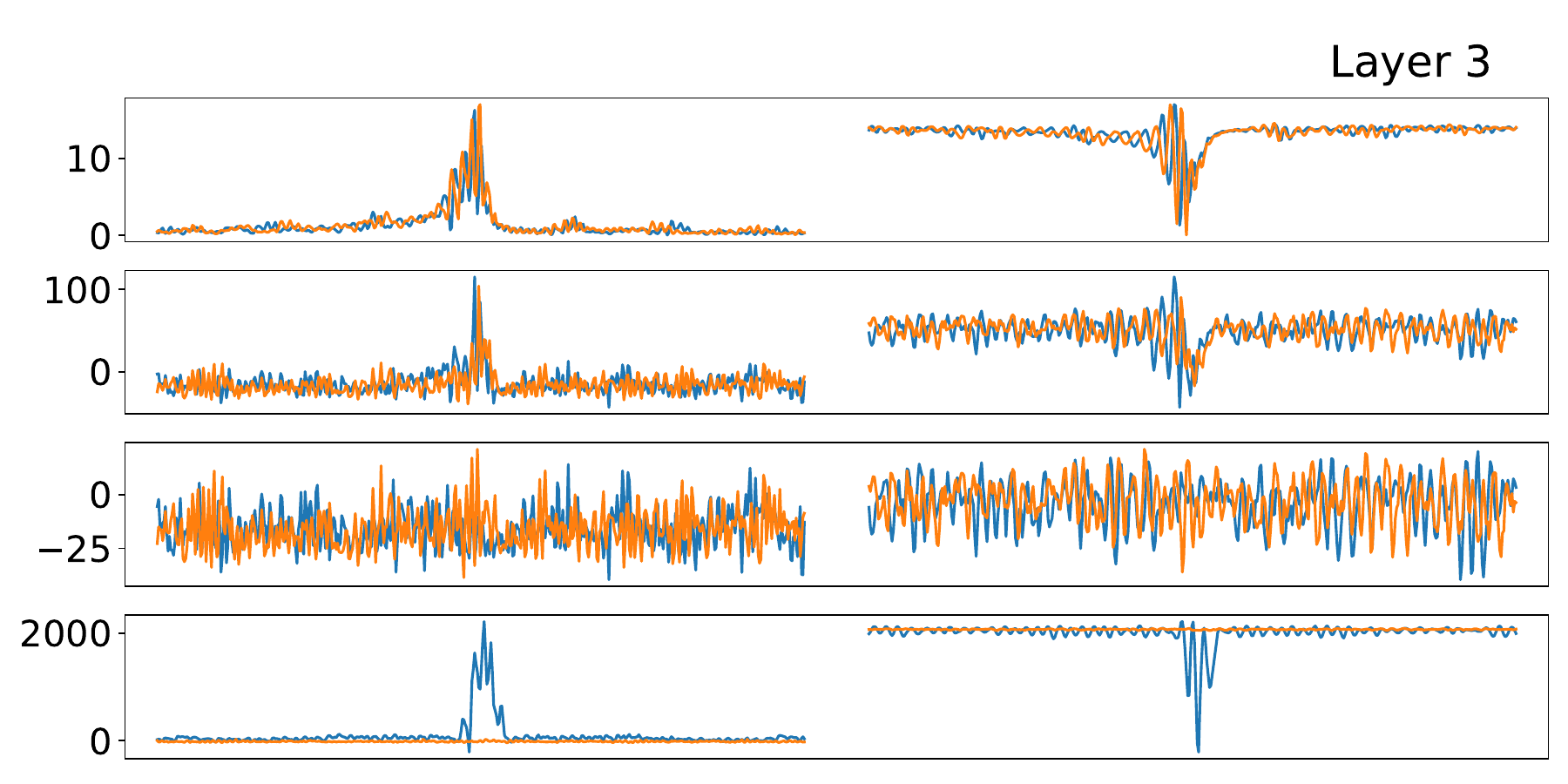}}%
    
    \vspace{1pt}\par

    \leavevmode
    \raisebox{0pt}{\includegraphics[width=0.48\linewidth]{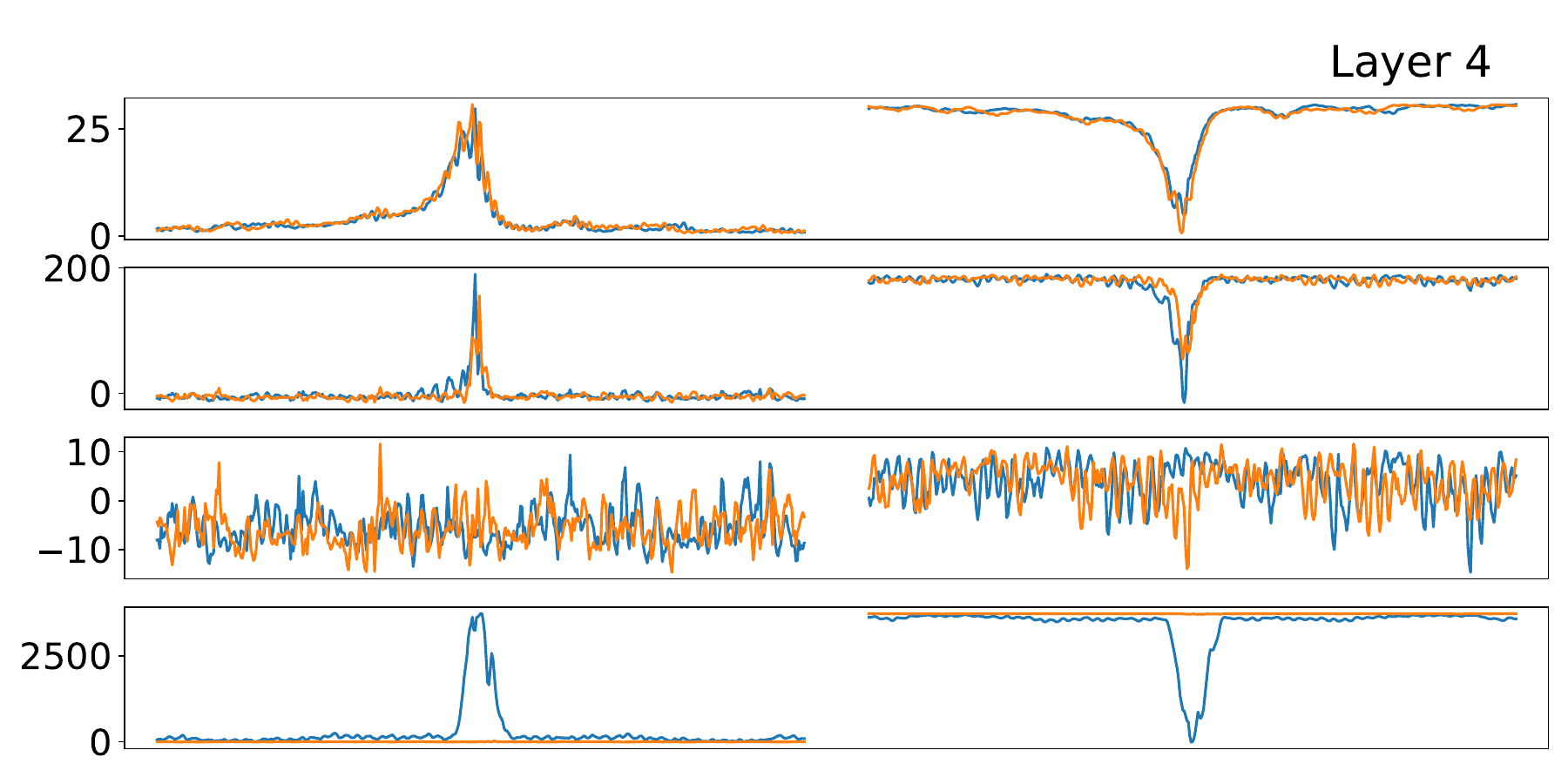}}%
    \hfill
    \raisebox{0pt}{\includegraphics[width=0.48\linewidth]{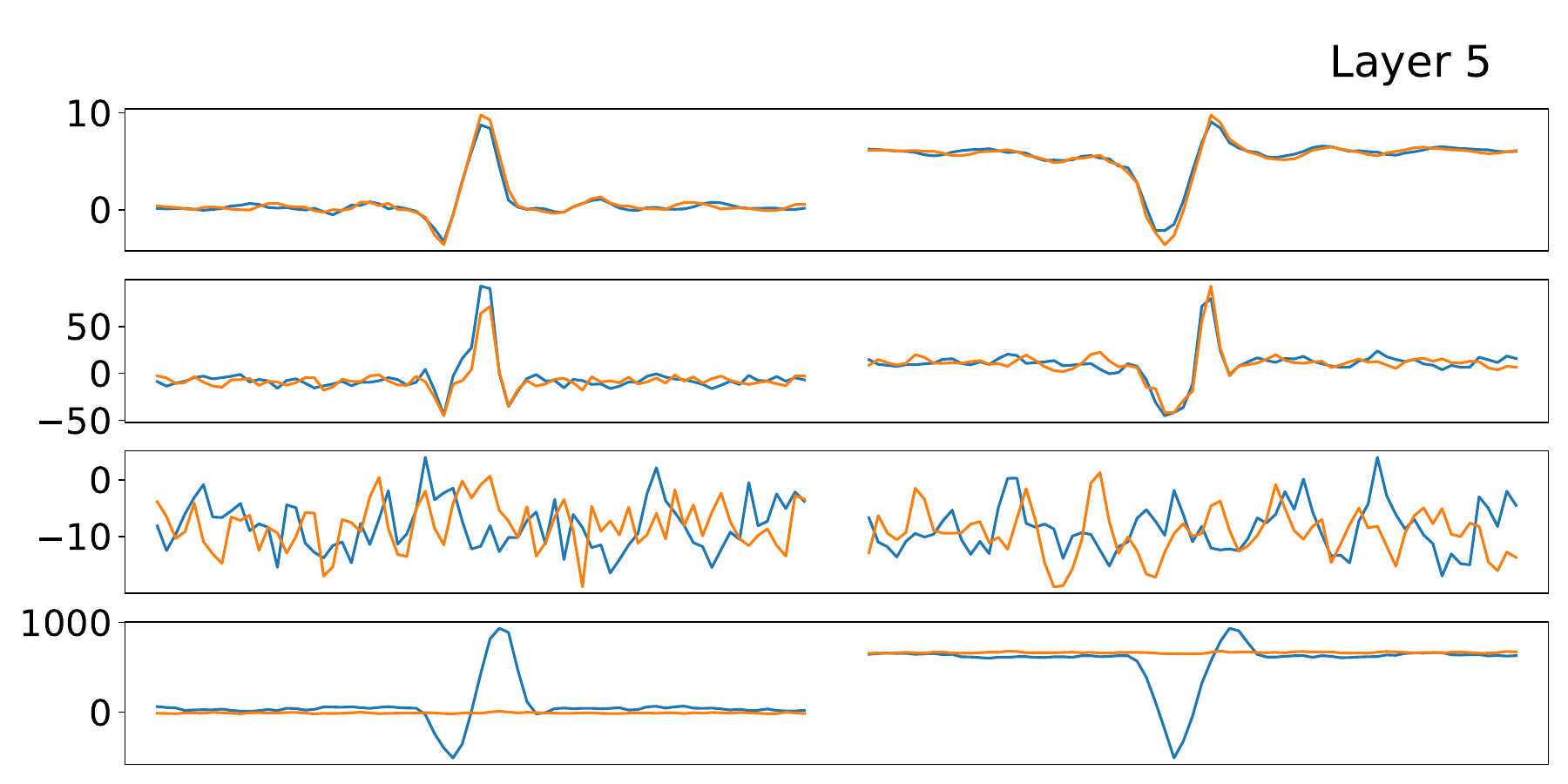}}%
    
    \vspace{1pt}\par

    \leavevmode
    \raisebox{0pt}{\includegraphics[width=0.48\linewidth]{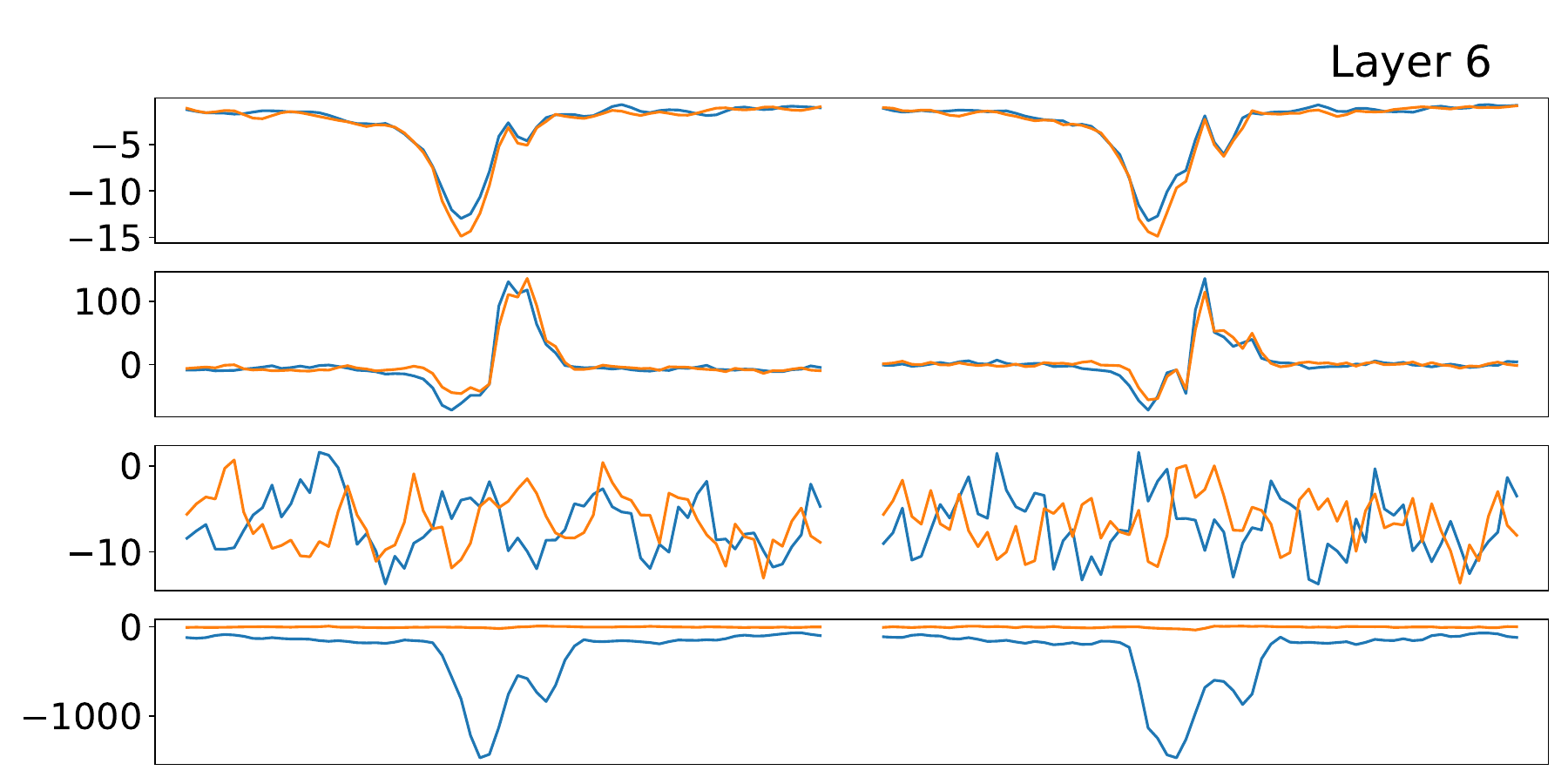}}%
    \hfill
    \raisebox{0pt}{\includegraphics[width=0.48\linewidth]{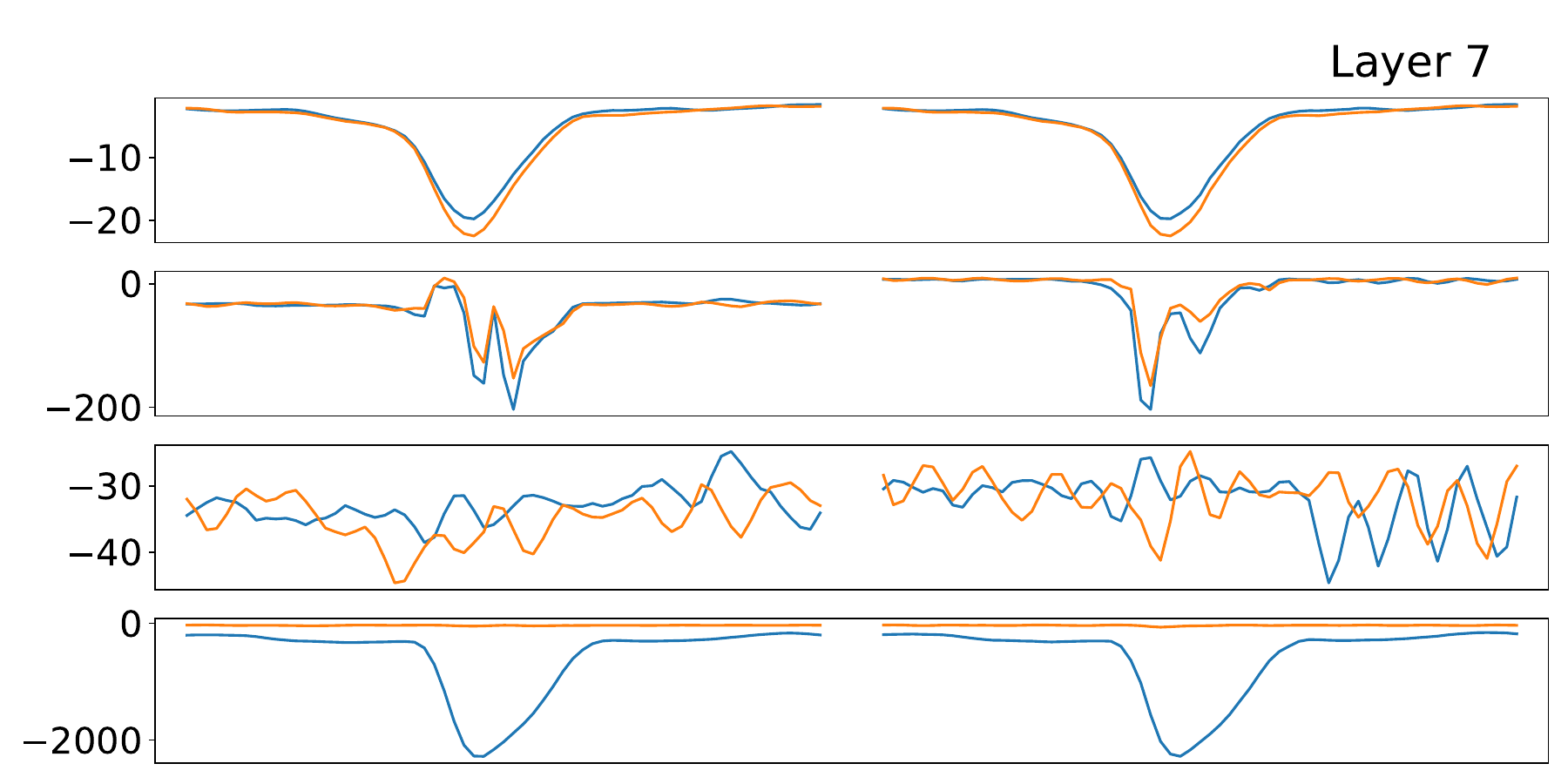}}%

    \caption{The input data and the first two feature maps from the convolutional layers. The input contains four types of data: the first is simulated GW waveform, the second is synthetic GW signal, the third is pure noise, and the fourth is glitch. The blue and orange lines in input represent the H1 detector and the L1 detector, respectively. The feature maps in each convolutional layer correspond to the four types of data.}
    \label{FIG.3}
\end{figure}

Figure~\ref{FIG.3} shows the four samples and their feature maps of various layers. 
In the input layer, blue and orange lines represent the H1 and L1 strain data, respectively. The same color scheme is used to differentiate between H1- and L1- feature representations in all feature maps. 
We observed that the H1-feature and L1-feature in the feature maps of the waveform and GW signal exhibit some sharp peaks and valleys, particularly as the depth of the layers increases. This phenomenon may imply that the baseline-CNN has extracted information indicating the transient nature of GW sources.
With the deepening of the convolution layers, the shapes of the H1-feature and L1-feature in the feature maps of GW signal become increasingly similar to those of the waveform. Furthermore, the H1-feature and L1-feature extracted from both the waveform and the GW signal show a close match, indicating that the network is effectively capturing the underlying signal characteristics. 
In contrast,  the H1-feature and L1-feature in the feature maps of the noise present high-frequency oscillation characteristics, with no significant matching between them.
The glitches shows markedly different behavior. The H1- and L1-features differ significantly: the H1-feature exhibits sharp variations with strong peaks and valleys, while the L1-feature appears relatively smooth. Interestingly, the H1-features exhibit a strong resemblance to those of the waveform.

\begin{figure}[htpb]
    \centering
     \includegraphics[width=1\linewidth]{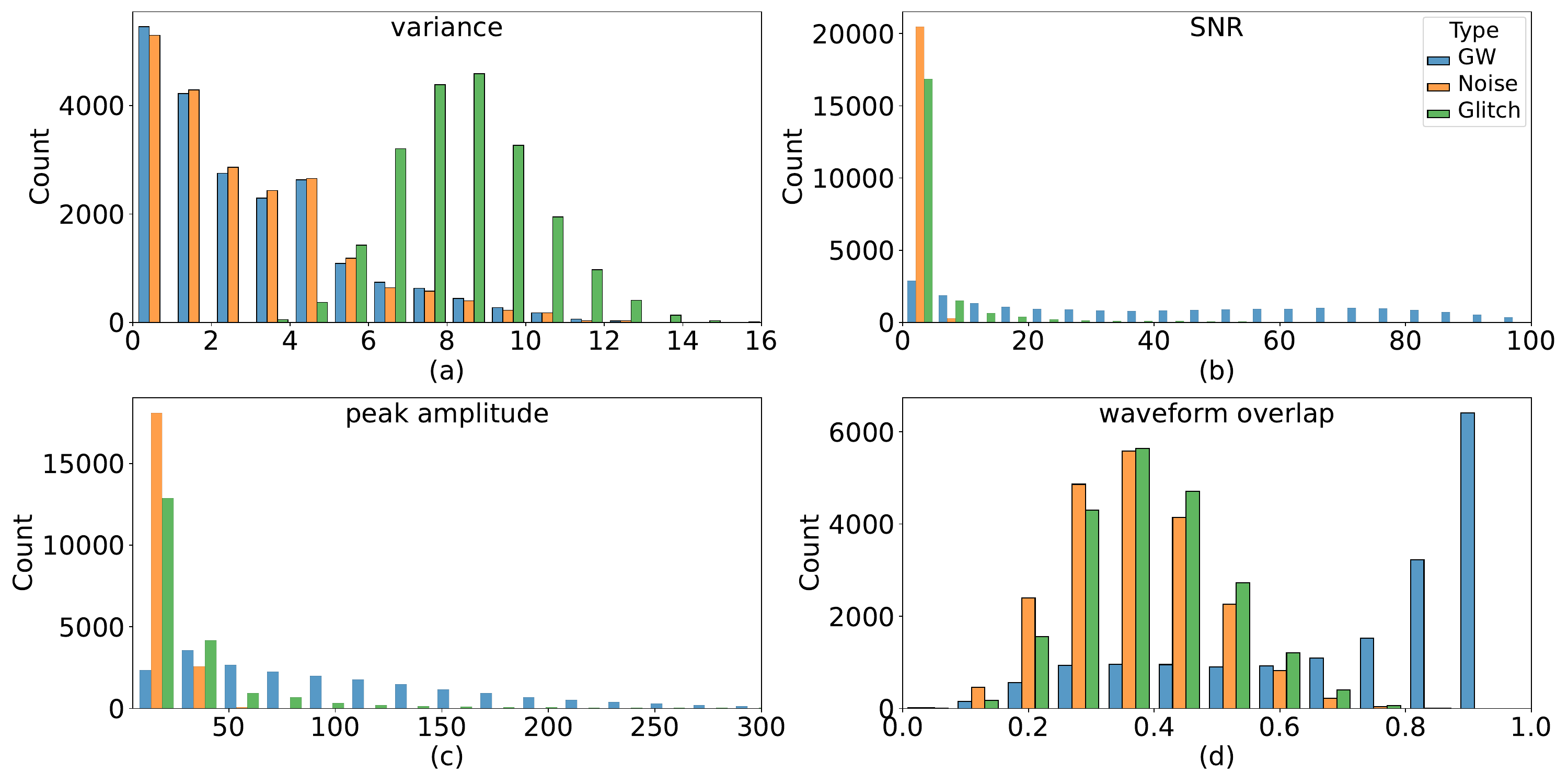}
    \caption{The histogram of the four extracted features across three types of data is presented in four panels labeled (a), (b), (c), and (d). Each panel represents the frequency distribution of one of the extracted features for GW signals (blue), Noise (orange), and Glitches (green).}
    \label{FIG.4}
\end{figure}

\subsection{Features Analysis}\label{5.2}

In Section~\ref{3.3}, we discussed how to extract the four handcrafted features. Figure~\ref{FIG.4} shows the distribution of four handcrafted features across three types of data in the training set of Dataset 2. In each panel, two types of data exhibit similar distributions, while the other shows a significant difference. In panel (a), the $\mathbf{variance}$ of the GW signals and noise exhibits a similar distribution, mainly concentrated below 6, while the $\mathbf{variance}$ of glitches is primarily distributed around 9. The $\mathbf{variance}$ of glitches is significantly higher than that of other types, indicating a significant difference between the H1-feature and L1-feature in the feature maps. This phenomenon is straightforward to understand, as the likelihood of glitches occurring simultaneously in both detectors is very low. Consequently, the features extracted by the baseline-CNN from the strain data of the two detectors exhibit significant differences.
In panels (b) and (d), the $\mathbf{SNR}$ and $\mathbf{waveform\ overlap}$ of GW signals are significantly higher, indicating that the H1-feature and L1-feature have a good match.
It shows that for the same GW event detected by different detectors, the features extracted by the baseline-CNN are similar. 
In panel (c), the $\mathbf{peak\ amplitude}$ of noise and glitches are mainly distributed in a low range, whereas the $\mathbf{peak\ amplitude}$ of GW signals are relatively uniformly distributed.
Variations in handcrafted features across different types of data validate the effectiveness of our manually extracted features.

\begin{table}[htb]
\caption{Feature Importance Ranking from the RF Classifier.}
\centering
\begin{tabular}{ccc}
\toprule
Rank & Feature  & Feature Importance \\
\midrule
1 & $\mathbf{variance}$  & 0.3943 \\
2 & $\mathbf{CNN\ probability}$    & 0.2583 \\
3 & $\mathbf{SNR}$       & 0.1629 \\
4 & $\mathbf{waveform\ overlap}$ & 0.1219 \\
5 & $\mathbf{peak\ amplitude}$   & 0.0626 \\
\bottomrule
\end{tabular}
\label{tab:feature_importance}
\end{table}

Table~\ref{tab:feature_importance} shows the ranking of feature importance of the RF classifier. A higher feature importance indicates a greater contribution to the model’s decision-making process. 
The feature importance of $\mathbf{CNN\ probability}$ is 0.2583, while the combined feature importance of the handcrafted features ($\mathbf{variance}$, $\mathbf{SNR}$, $\mathbf{waveform\ overlap}$, $\mathbf{peak\ amplitude}$) is 0.7417, indicating that the extracted features dominate the decision-making process.
Specifically, $\mathbf{variance}$ holds the highest feature importance (0.3943), which directly correlates with the fact that the glitch distribution is clearly separated from those of GW signals and noise in Figure~\ref{FIG.4}(a). While the $\mathbf{variance}$ of GW signals and noise are mostly concentrated below 6, glitches exhibit much higher $\mathbf{variance}$, typically centered around 9. This large discrepancy highlights the distinct cross-detector inconsistency introduced by glitches, making variance a key discriminative feature for classification.
The $\mathbf{CNN\ probability}$ ranks second with a feature importance of 0.2583, followed by $\mathbf{SNR}$ (0.1629) and $\mathbf{waveform\ overlap}$ (0.1219). The $\mathbf{peak\ amplitude}$ feature has the lowest feature importance of 0.0626, suggesting that it has a relatively smaller influence on the model’s classification.
The feature importance analysis confirms that our handcrafted features are well-aligned with the underlying data distribution, enabling the classifier to distinguish GW signals from noise and glitches effectively.

From the above discussion, it is evident that although the baseline-CNN was trained solely on GW signals and noise, the network also captured characteristics of glitches. This observation suggests that the pre-trained CNN serves as an effective feature extractor. By modifying or retraining only the classifier on top of this pre-trained network, it is possible to address a broader range of challenges more effectively.
This approach leverages the robust feature extraction capabilities developed during the initial training phase, allowing for efficient adaptation to new tasks or datasets by focusing adjustments on the classification layers. Such a strategy not only economizes on computational resources but also enhances the versatility of the model across various applications.

\subsection{Performance}\label{5.3}

\begin{figure}[htp]
    \centering
    \includegraphics[width=0.8\linewidth]{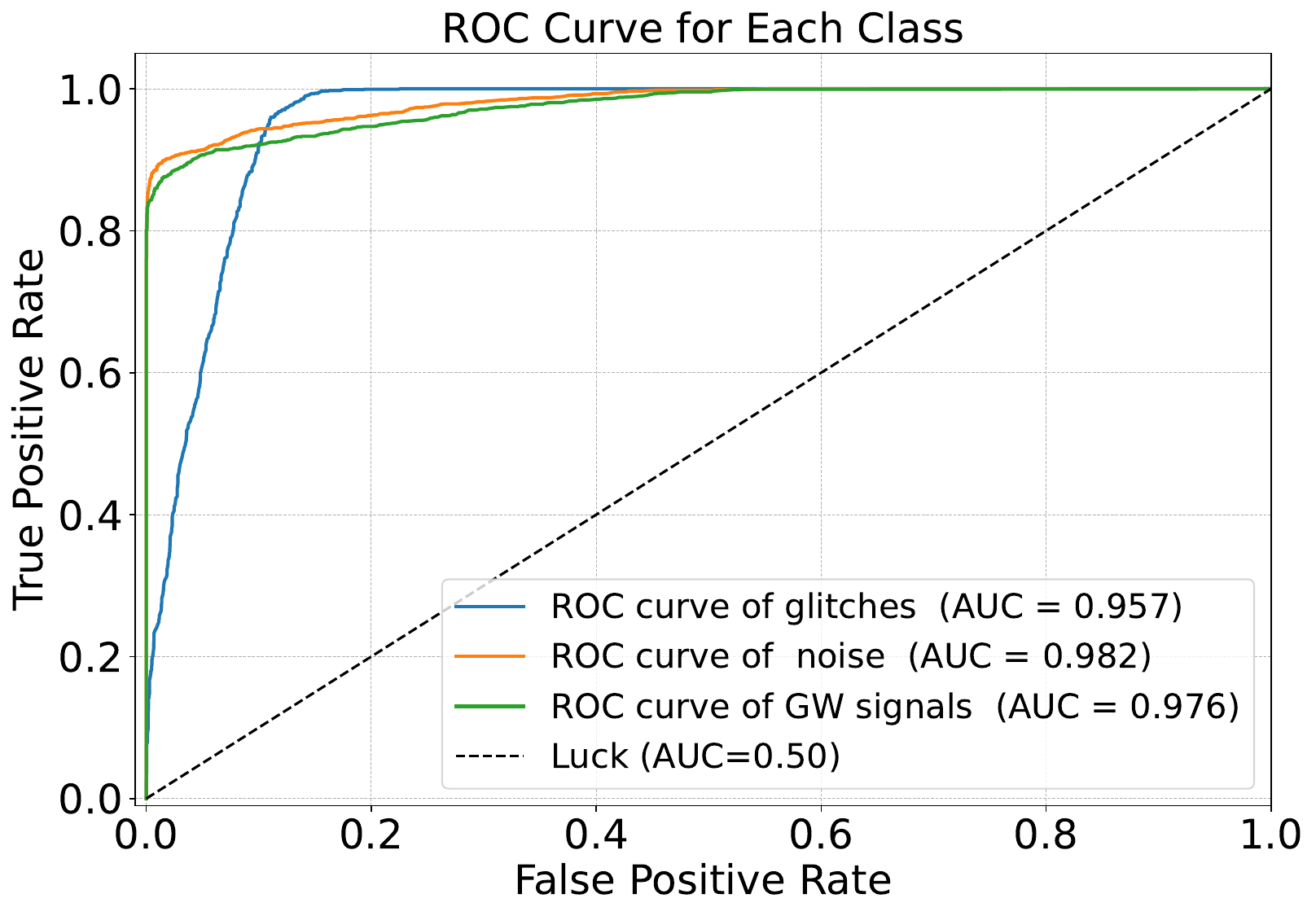}
    \caption{The ROC curves corresponding to each class in the CNN-RF model. The blue curve corresponds to GW signals, the orange to noise, and the green to glitches. The black curve labeled ‘luck’ represents the ROC curve of a random classifier (i.e., random guessing), which yields an expected AUC of 0.5.}
    \label{FIG.5}
\end{figure} 

In this subsection, we evaluate the performance of the CNN-RF model for each class. The Receiver Operating Characteristic (ROC) curve is commonly used to assess binary classification performance, where a curve closer to the upper-left corner and a larger area under the curve (AUC) indicate better performance~\cite{FAWCETT2006861}. For multi-class scenarios, we employ the one-vs-rest (OvR) strategy, treating each class as the positive class and aggregating the remaining classes as the negative class to construct individual ROC curves~\cite{Jin2021}. This process is repeated sequentially for all other categories using the same methodology.

Figure~\ref{FIG.5} presents the ROC curves and the corresponding AUC values for each class in the CNN-RF model. The ROC curves for GW signals and noise are highly similar, indicating that the model achieves similar performance for both classes. In contrast, the ROC curve for glitches shows a slightly different pattern, reflecting the model’s ability to distinguish glitches more distinctly from the other classes. The AUC values for noise and GW signals are 0.982 and 0.976, respectively, indicating the model’s high performance in classifying these two classes. Although the AUC for glitches is slightly lower at 0.957, it still reflects a strong classification capability. Overall, the consistently high AUC values ($\ge 0.95$) suggest that the CNN-RF model achieves robust performance across all categories.

Notably, while the feature extractor (baseline-CNN) in the CNN-RF model was originally developed based on GW signals and noise, its learned representations also enable effective classification of glitches when combined with a RF classifier. This highlights the CNN's capacity to generalize features across different signal types, even beyond its original training distribution.

\subsection{Ablation Study of Feature}\label{5.4}

\begin{table}[htbp]
\centering
\caption{AUC scores of different feature configurations across three classes.}
\begin{tabular}{lccc}
\toprule
\textbf{Feature} & \textbf{GW signals} & \textbf{Noise} & \textbf{Glitches} \\
\midrule
Full           & $\mathbf{0.976}$ & $\mathbf{0.982}$ & $\mathbf{0.957}$ \\
$\mathbf{CNN\ probability}$ only              & 0.957 & 0.809 & 0.753 \\
Handcrafted features only    & 0.974 & $\mathbf{0.982}$ & 0.955 \\
w/o $\mathbf{variance}$ + $\mathbf{peak\ amplitude}$      & 0.971 & 0.821 & 0.783 \\
w/o $\mathbf{SNR}$ + $\mathbf{waveform\ overlap}$       & 0.964 & 0.975 & 0.950 \\
\bottomrule
\end{tabular}
\medskip

\begin{minipage}{0.85\linewidth}
\footnotesize
Note: “w/o” stands for “without”, indicating the features that were excluded in specific configurations.
\end{minipage}

\label{tab:feature_ablation_auc}
\end{table}

To evaluate the individual and collective contributions of handcrafted features to classification performance, we conducted a feature ablation study. Different subsets of features were selectively removed, and the RF classifier was retrained to assess changes in performance. The AUC was used as the evaluation metric across the three classes, with results summarized in Table~\ref{tab:feature_ablation_auc}.

For clarity, the handcrafted features were grouped into two functional categories: ($\mathbf{SNR}$ and $\mathbf{waveform\ overlap}$), which reflect waveform similarity and inter-detector coherence; and ($\mathbf{variance}$ and $\mathbf{peak\ amplitude}$), which capture statistical properties of the signal distribution. We conducted group-wise ablation by removing each pair and observing the performance impact.
The full model, incorporating both the $\mathbf{CNN\ probability}$ and all four handcrafted features, achieves the best performance across all classes. Removing the $\mathbf{CNN\ probability}$ and using only handcrafted features leads to only a marginal performance drop, indicating that the handcrafted features alone are highly informative and play a central role in classification.
In contrast, using only the $\mathbf{CNN\ probability}$ causes a significant decrease in AUC, especially for the noise and glitch classes (dropping from 0.982 to 0.809, and from 0.957 to 0.753, respectively). This suggests that while the baseline-CNN is effective at identifying GW signals, it lacks robustness against non-Gaussian artifacts without additional handcrafted cues. This limitation is expected, as the baseline-CNN was originally trained solely for binary classification between GW signals and noise.
The group-wise ablation further highlights the relative importance of different feature types. Removing $\mathbf{variance}$ and $\mathbf{peak\ amplitude}$ results in a sharp performance decline for the noise and glitch classes, with AUCs falling from 0.982 to 0.821 and from 0.957 to 0.783, respectively. This underscores their effectiveness in capturing amplitude-based differences between clean data and transient disturbances. Conversely, removing $\mathbf{SNR}$ and $\mathbf{waveform\ overlap}$ has a milder impact, primarily reducing the AUC for the GW class (from 0.976 to 0.964), suggesting that these features are more relevant for identifying coherent astrophysical signals.

Overall, the results confirm the complementary nature of the handcrafted features and their crucial role in enhancing the model’s ability to differentiate between signal types, especially in the presence of noise and glitches.

\subsection{Sensitivity}\label{5.5}

\begin{figure}[htp]
    \centering
    \includegraphics[width=0.8\linewidth]{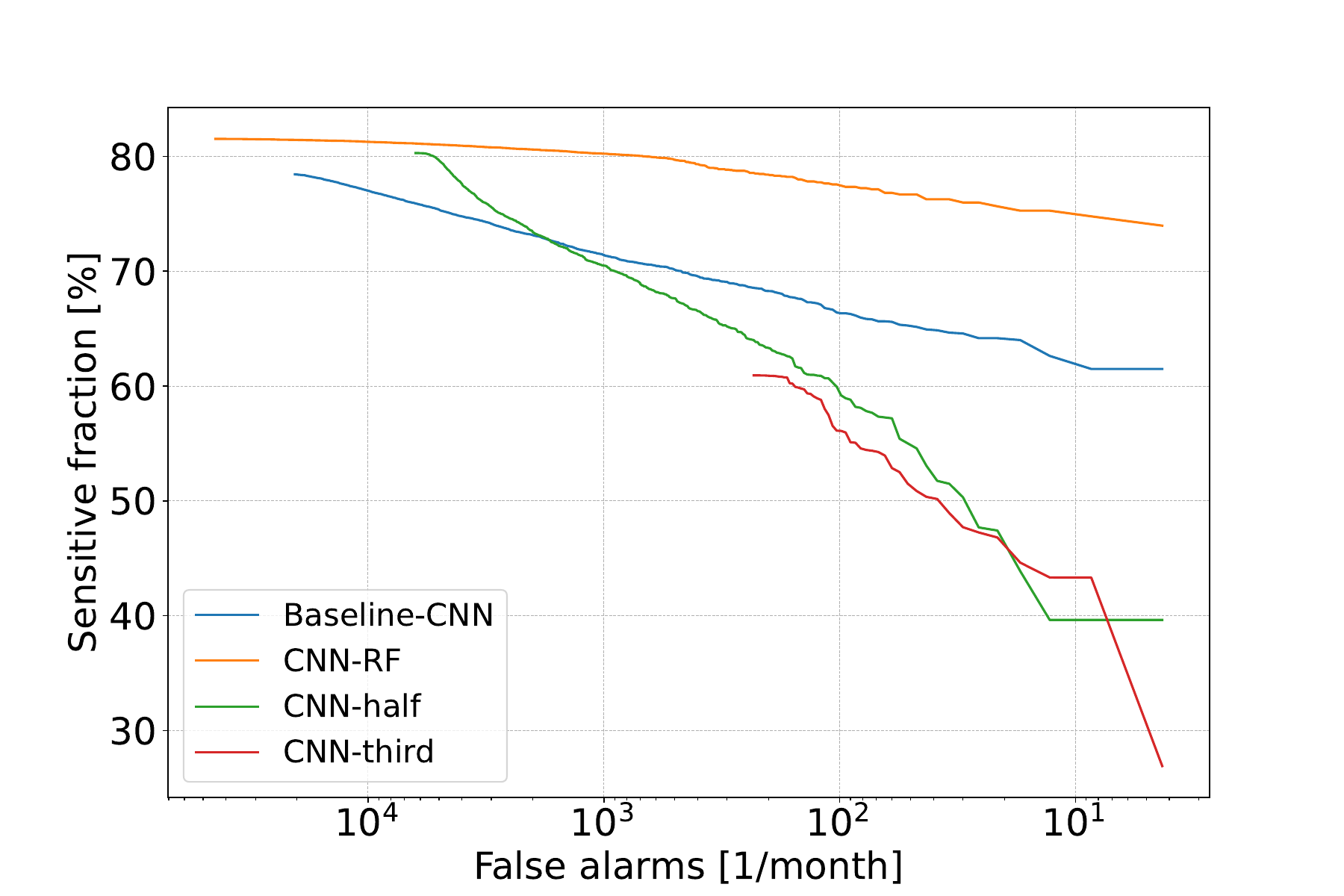}
    \caption{The sensitive fraction of models at various FAR. The orange, blue, green, and red curves correspond to the CNN-RF, baseline-CNN, CNN-half, and CNN-third models, respectively.}
    \label{FIG.6}
\end{figure}

\begin{figure}[htp]
    \centering
    \includegraphics[width=0.8\linewidth]{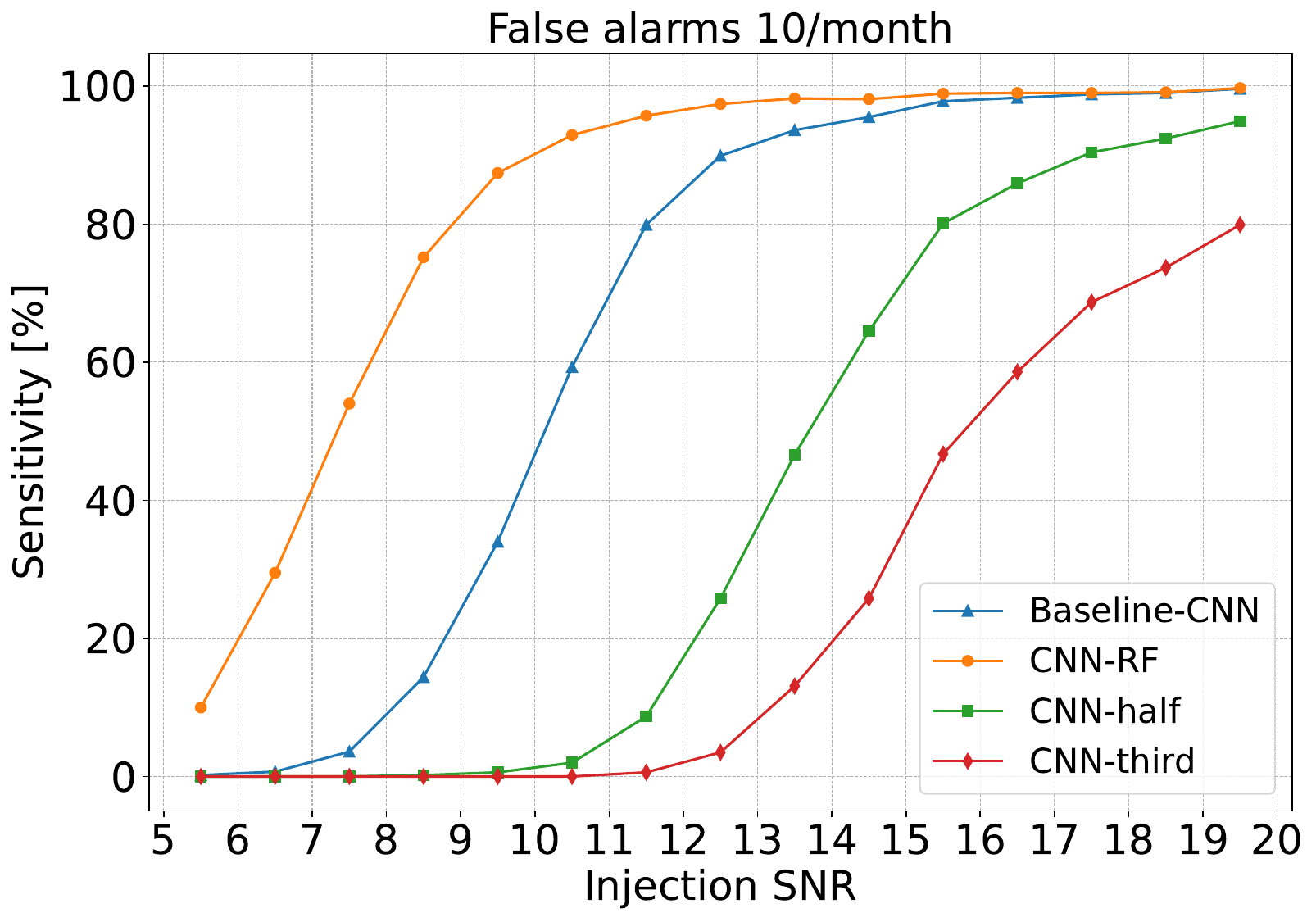}
    \caption{The line chart illustrates the sensitivity distribution of models at a FAR of 10 per month. The blue upward triangle (baseline-CNN), orange circle (CNN-RF), green square (CNN-half), and red diamond (CNN-third) represent the respective models in the figure. The $x$-axis denotes the injection SNR, indicating the SNR of the GW signals at the time of injection, while the $y$-axis shows sensitivity, the proportion of detected signals. }
    \label{FIG.7}
\end{figure}

In this subsection, we discuss the sensitivities of various models.  
When training the CNN-RF model, we incorporated an additional data type (glitches) beyond what was used in the baseline-CNN.  
To enable a more equitable comparison, we designed two additional CNN architectures: CNN-half and CNN-third. Both share the same core architecture as the baseline-CNN but differ in their training configurations. 
The CNN-half model uses the same architecture as the baseline-CNN. Its training dataset consists of 50\% GW signals and 50\% non-GW signals, with the latter evenly split between pure noise and glitches (each contributing 25\% of the total dataset).  
In contrast, the CNN-third model extends the baseline-CNN into a three-class classification task. Its training data are uniformly distributed among three categories: pure noise (33.3\%), GW signals (33.3\%), and glitches (33.3\%). Apart from modifying the output layer to accommodate three classes, the CNN-third model retains the same architecture as the baseline-CNN. It is important to note that the total number of training samples for both CNN-half and CNN-third is identical to that of the baseline-CNN.

Figure~\ref{FIG.6} presents a comparison of the sensitive fractions achieved by four different models across a range of false alarm rates. The sensitive fraction represents the proportion of correctly detected GW signals, serving as a key indicator of detection performance under specific false alarm constraints. 
The CNN-RF consistently demonstrates the highest sensitive fraction across the entire false alarm rate spectrum, indicating that combining convolutional features with a RF classifier significantly boosts detection sensitivity. This performance gain is attributed to the hybrid model’s ability to leverage both learned features and decision-level fusion.
The baseline-CNN ranks second in overall performance, outperforming both CNN-half and CNN-third in most cases. Although the CNN-half model exhibits improved sensitivity in the mid-to-high false alarm rate regime, its performance drops significantly in the low false alarm rate region compared to the baseline-CNN.
The CNN-third model performs the worst among the four, indicating that the added complexity of three-class classification may introduce more confusion and reduce overall sensitivity.
In practical GW detection scenarios, performance at low false alarm rates is particularly critical. As summarized in Table~\ref{tab:Sensitive_fraction}, the CNN-RF model achieves the highest sensitive fraction across all low FAR levels, with a relative improvement of over 21\% compared to the baseline-CNN at a FAR of 10 per month. Both CNN-half and CNN-third exhibit significantly reduced performance under strict false alarm constraints.

\begin{table}[htpb]
\caption{Sensitive fraction of models at different FAR per month.}
\centering
\begin{tabular}{ccccl}
\toprule
& \multicolumn{3}{c}{FAR} & \\
\cmidrule(lr){2-4}
\textbf{Model} & 10 & 100 & 1000 \\
\midrule
Baseline-CNN & 62$\%$ & 66$\%$ & 71$\%$ \\
CNN-RF & 75$\%$ & 78$\%$ & 80$\%$ \\
CNN-half & 40$\%$ & 59$\%$ & 71$\%$ \\
CNN-third & 43$\%$ & 56$\%$ & 61$\%$ \\
\bottomrule
\end{tabular}
\label{tab:Sensitive_fraction} 
\end{table}

The sensitive fraction results indicate that the CNN-RF model detects more GW events than the other models at the same FAR. To further understand this improvement, we analyze the distribution of injected SNRs for events detected by various models.
Figure~\ref{FIG.7} shows the sensitive distribution as a function of the injected SNRs for detected events across different models at a FAR of 10 per month. 
As the injection SNR increases, all models exhibit an upward trend in the number of detected GW events. However, the CNN-RF model consistently outperforms the others, reaching over 90\% sensitivity at an injection SNR in the range of 10 to 11—earlier than any other model. In contrast, the baseline-CNN still lags by approximately 30\% in this range, highlighting the CNN-RF’s superior ability to detect signals even at relatively low SNRs.
Remarkably, the CNN-RF model maintains a sensitivity of 10\% at the lowest tested SNR, whereas other models approach 0\% under the same conditions. CNN-half and CNN-third, in particular, exhibit significantly reduced detection capabilities, with near-zero sensitivity for SNRs below 10.
Taken together with the sensitive fraction analysis, these findings confirm that the enhanced performance of the CNN-RF model is primarily attributed to its improved sensitivity in the low-SNR regime. This underscores its strong discriminative power and robustness in challenging detection scenarios.

\subsection{Handling of Glitch Features in Model Training}\label{5.6}

During the training phase, CNN-RF, CNN-half, and CNN-third were all trained on datasets containing glitches. This approach was adopted because real-world data often include numerous glitches that frequently trigger false alarms. By including glitches in the training process, we aimed to enhance the models’ ability to accurately distinguish glitches from GW signals, thereby minimizing the rate of false alarms. 

It is important to note that the four models (baseline-CNN, CNN-RF, CNN-half, and CNN-third) differ in how they learn and represent glitch-related features. The baseline-CNN was trained without exposure to glitches, and thus did not learn any explicit glitch representations. In contrast, CNN-half and CNN-third incorporated glitches with explicit labels, either as noise or as separate classes, leading them to learn glitch features under strong supervision. CNN-RF, on the other hand, employed a hybrid architecture: it first allowed the CNN to naturally extract abstract features from the data—including glitches—and then passed these representations to a RF, which was manually trained to discriminate between glitches and GW signals.

However, we observed that the sensitivities of CNN-half and CNN-third were lower than those of the baseline-CNN. This suggests that incorporating excessive glitches during training—especially under strong supervision—may compromise the ability of CNNs to detect GW signals. In contrast, CNN-RF demonstrated a sensitivity higher than that of the baseline-CNN, while maintaining robust classification of glitches. The enhanced robustness of CNN-RF likely comes from its hybrid architecture, which combines the CNN’s capacity for feature abstraction with the RF’s ability to mitigate overfitting and improve generalization.

\section{Conclusion}\label{6}

In this work, we propose a hybrid CNN-RF model for GW detection, designed to classify GW signals, noise, and glitches. The model integrates a pre-trained CNN-based feature extractor with a RF classifier, enhanced by four interpretable handcrafted features derived from the final convolutional layer of the CNN. These features ($\mathbf{SNR}$, $\mathbf{waveform\ overlap}$, $\mathbf{variance}$, and $\mathbf{peak\ amplitude}$) capture statistical and waveform coherence properties between dual-detector data, enabling the RF classifier to establish robust decision boundaries.

The key findings of this paper are summarized as follows:
\begin{enumerate}
    \item 
    The CNN serves as an effective feature extractor. Notably, although the baseline-CNN is trained exclusively on GW signals and noise, it is also capable of extracting useful information from glitches. This capability allows us to modify or retrain only the classifier layer on top of the pre-trained network for new tasks, thereby significantly reducing computational costs and training time while maintaining robust performance.

    \item 
    The handcrafted features demonstrate strong discriminative power across different data types. These features capture both statistical deviations and inter-detector matches, enabling the RF classifier to form more accurate decision boundaries. Feature importance analysis confirms that these handcrafted features dominate the decision-making process, highlighting their critical role in improving model performance.

    \item 
    The CNN-RF model demonstrates strong performance in processing long-duration data, achieving the highest sensitivity across all FAR levels compared to other models. At a FAR of 10 events per month, it achieves a relative improvement of over 21\% compared to the baseline-CNN and remains capable of detecting signals with the lowest injection SNR, whereas all other models fail to detect any signals under the same condition.

\end{enumerate}

This hybrid approach enhances both the interpretability and performance of CNN-based methods for GW detection by incorporating physically motivated features. Previous studies have shown that CNNs can reach sensitivity levels comparable to, and in some cases exceeding, those of matched filtering under controlled conditions~\cite{2018PhLB..778...64G,2018PhRvL.120n1103G,2023PhRvD.107b3021S}.
While many works have focused on refining network architectures and training strategies to further improve performance, the resulting gains in detection capability have often been relatively modest.
These observations suggest that future efforts could benefit from placing greater emphasis on improving internal model interpretability and feature transparency, rather than solely focusing on increasingly complex or advanced architectures.

Despite these promising results, several limitations remain. First, the inclusion of four handcrafted features and an additional RF classifier introduces increased computational overhead. On the one-week dataset, the hybrid model requires approximately four times more inference time than the baseline-CNN model, which may limit its applicability in real-time or low-latency detection pipelines. Second, the current study focuses exclusively on BBH signals using a two-detector configuration (H1 and L1). In contrast, real-world GW detection scenarios involve a broader range of CBC sources, including BNS and NSBH mergers, and multi-detector networks that may include Virgo, KAGRA, or LIGO-India~\cite{2023arXiv230107522U}. The extent to which the proposed features and classification framework generalize to these more complex observational settings remains an open question.

To address these limitations, we plan to extend the hybrid framework in several directions. Future work aims to accommodate a broader range of astrophysical sources, including unmodeled bursts from core-collapse supernovae and long-duration signals expected in space-based missions such as LISA~\cite{2017arXiv170200786A}. We also intend to incorporate data from additional detectors (e.g., Virgo, KAGRA, LIGO-India) for joint network analysis, thereby enhancing robustness and generalizability. Furthermore, we will explore more flexible and data-driven feature selection strategies that retain physical interpretability, with the goal of improving both the sensitivity and adaptability of the model across various gravitational-wave detection tasks.

\section*{Data availability statement}

The original data are provided by the LIGO collaboration and can be accessed at \url{https://www.gw-openscience.org}. The data processing pipeline is available via the corresponding open-source repository.

\ack
We thank the Gravitational Wave Open Science Center (GWOSC) for providing access to detector data and analysis tools. 
This work was supported by The Central Guidance on Local Science and Technology Development Fund of Sichuan Province (2024ZYD0075); National Natural Science Foundation of China Grant Nos. 12105032; the Natural Science Foundation of Chongqing 18 No. cstc2021jcyj-msxmX0481. H.W. is partially supported by the National Key Research and Development Program of China (Grant No. 2021YFC2203004) and National Science Foundation of China (NSFC) under Grant No. (12405076).

\section*{References}
\bibliographystyle{unsrt}
\bibliography{references}

\end{document}